%% file: main.tex
\documentclass[preprint,12pt]{elsarticle}
\usepackage[labelsep=period]{caption}
\usepackage{graphicx} 
\graphicspath{ {./images/} }
\usepackage{amsmath}
\usepackage{amssymb}
\usepackage{setspace}
\usepackage{hyperref}
\usepackage[margin=2.5cm]{geometry}
\usepackage{algorithm}
\usepackage{algpseudocode}
\usepackage{xcolor}
\usepackage{todonotes}
\usepackage{multirow} 
\usepackage{array}
\newcolumntype{C}[1]{>{\centering\arraybackslash}p{#1}}

\input{macros}


\begin{document}

\begin{frontmatter}

\title{Amortized Variational Inference for Joint Posterior and Predictive Distributions in Bayesian Uncertainty Quantification}

\author{Nan Feng\textsuperscript{a} and Xun Huan\textsuperscript{a}}

\date{July 2024}

\affiliation[a]{organization={Department of Mechanical Engineering, University of Michigan},
            city={Ann Arbor}, 
            state={Michigan},
            country={United States}}

\begin{abstract}

Bayesian predictive inference propagates parameter uncertainty to quantities of interest through the posterior-predictive distribution. In practice, this is typically performed using a two-stage procedure: first approximating the posterior distribution of model parameters, and then propagating posterior samples through the predictive model via Monte Carlo simulation. This sequential workflow can be computationally demanding, particularly for high-fidelity models such as those governed by partial differential equations.

We propose a variational Bayesian framework that directly targets the posterior-predictive distribution and jointly learns variational approximations of both the posterior and the corresponding predictive distribution. The formulation introduces a variational upper bound on the Kullback--Leibler divergence together with moment-based regularization terms. The variational distributions are trained in an amortized manner, shifting computational effort to an offline stage and enabling efficient online inference. Numerical experiments ranging from analytical benchmarks to a finite-element solid mechanics problem demonstrate that the proposed method achieves more accurate predictive distributions than conventional two-stage variational inference, while substantially reducing the cost of online predictive inference.
\end{abstract}


\end{frontmatter}

\section{Introduction}

High-fidelity computational models, often governed by partial differential equations (PDEs), play a central role in scientific and engineering analysis, prediction, and design across disciplines such as structural mechanics, fluid dynamics, and materials science \citep{hughes2012finite, ferziger2019computational, de2011computational}. While these models have become increasingly sophisticated, single deterministic simulations are generally insufficient to support reliable decision-making. Real-world systems are inevitably affected by uncertainty arising from stochastic environments, variability in material properties, manufacturing tolerances, and incomplete or noisy measurements. Neglecting such uncertainties can lead to misleading predictions and potentially unsafe designs or decisions.

These challenges have motivated the development of uncertainty quantification (UQ)~\citep{smith2013uncertainty, soize2017uncertainty, Various2017}, which seeks to represent, propagate, and reduce uncertainty in computational models. A central task in UQ is \emph{inverse UQ}, which aims to infer uncertain model parameters from observational data that are often sparse and noisy. Within this setting, the Bayesian paradigm~\citep{gelman2013bayesian, sivia2006data} provides a principled probabilistic framework for updating prior knowledge of uncertain parameters, denoted $\theta$, in light of observed data $y$. Bayes’ theorem defines the posterior distribution\footnote{For clarity of presentation, we restrict attention to continuous random variables with associated probability densities; analogous formulations apply to discrete random variables with probability mass functions.} as
\begin{align*}
p(\theta | y) = \frac{p(y | \theta) p(\theta)}{p(y)},
\end{align*}
where $p(\theta)$ denotes the prior distribution, $p(y | \theta)$ is the likelihood whose evaluation typically requires solving a forward model, and $p(y)$ is the marginal likelihood (model evidence). The objective of inverse UQ is therefore to characterize the posterior distribution $p(\theta | y)$. For nonlinear models, this posterior distribution is generally analytically intractable.

Markov chain Monte Carlo (MCMC) methods~\citep{brooks2011handbook} have long been the dominant numerical tools for approximating Bayesian posteriors. Modern variants such as Hamiltonian Monte Carlo and the No-U-Turn Sampler exploit gradient information to improve sampling efficiency~\citep{duane1987hybrid, neal2011mcmc, betancourt2017conceptual, hoffman2014no}. Although MCMC methods are asymptotically exact, their practical use is often limited by slow convergence, challenging diagnostics, and high computational cost, particularly when each likelihood evaluation requires solving a high-fidelity PDE-based model. To address these scalability challenges, variational inference (VI) has emerged as a compelling alternative \citep{jordan1999introduction, blei2017variational, zhang2018advances}. VI reformulates Bayesian inference as an optimization problem by approximating the posterior distribution $p(\theta | y)$ with a tractable, parameterized family of distributions $q_{\lambda}(\theta|y)$. Common choices include Gaussian variational families, such as mean-field or structured Gaussian approximations, as well as particle-based representations obtained through Stein variational methods~\citep{Liu2016, Detommaso2018, Chen2020}.

In many scientific and engineering applications, however, inference of the parameters $\theta$ is not the ultimate objective. Instead, the primary goal is prediction: after inferring $\theta$, one seeks to quantify the resulting uncertainty in unobserved quantities of interest (QoIs), denoted $z$. The Bayesian framework addresses this objective through the posterior-predictive distribution
\begin{align*}
p(z | y) = \int p(z | \theta) p(\theta | y) \, \mathrm{d}\theta,
\end{align*}
which naturally accounts for uncertainty in the inferred parameters. In practice, approximating this integral typically relies on Monte Carlo sampling~\citep{robert2004monte,rubinstein2016simulation} from the posterior distribution, followed by forward propagation through the predictive model.

As a result, predictive inference in Bayesian UQ is typically performed through a two-stage procedure:
\begin{enumerate}
    \item an inverse problem is first solved to approximate the posterior distribution of the model parameters and generate posterior samples, and
    \item these samples are then propagated through the predictive model via an additional, often comparably expensive, Monte Carlo procedure to obtain samples from the posterior-predictive distribution.
\end{enumerate}
This sequential separation between posterior inference and predictive uncertainty propagation constitutes a major computational bottleneck in large-scale Bayesian UQ. Moreover, both stages can only be performed after the observational data $y$ become available, which can hinder rapid inference and prediction in time-sensitive settings such as real-time or online decision-making.

In this work, we introduce a variational Bayesian framework that circumvents the conventional two-stage procedure by directly targeting the posterior-predictive distribution. Rather than treating prediction as a post-processing step, the proposed approach formulates a single variational optimization problem that jointly learns approximations to both the posterior and the corresponding predictive distribution. Importantly, the variational distributions are constructed offline in an amortized manner, prior to observing any data $y$, allowing the computationally intensive training phase to be performed ahead of deployment. During online operation, newly observed data can be incorporated by simply evaluating the learned distributions at $y$, yielding the corresponding variational distributions with minimal additional cost.

The main contributions of this work are summarized as follows.
\begin{itemize}
    \item We propose a composite loss function that combines a variational upper bound on the Kullback--Leibler divergence between approximate and true posterior-predictive distributions with moment-based regularization that encourages agreement in key statistics (e.g., mean and variance).

    \item We employ deep neural networks to parameterize flexible variational distributions for both the posterior and posterior-predictive distributions, enabling amortized inference in which the learned distributions can be evaluated for new observations $y$ without retraining. 

    \item We validate the proposed framework through a hierarchy of numerical examples, ranging from low-dimensional linear and nonlinear benchmarks to a PDE-governed continuum solid mechanics problem. 
    In the numerical examples considered, the proposed method yields more accurate posterior-predictive estimates than the conventional approach, while substantially reducing the cost of online predictive inference.
\end{itemize}

The remainder of this paper is organized as follows. Section~\ref{s:background} introduces the Bayesian predictive formulation and summarizes the conventional two-stage posterior-predictive procedure under VI. Section~\ref{proposed_vimethod} presents the proposed joint variational formulation and the resulting algorithm. Section~\ref{numerical_examples} provides numerical results and comparative studies. Finally, Section~\ref{conclusions} summarizes the main findings and discusses directions for future research.
\section{Background} 
\label{s:background}

\subsection{Problem formulation}

Consider an observation forward model $G : \mathbb{R}^{d_{\theta}} \to \mathbb{R}^{d_y}$ that maps model parameters $\theta \in \mathbb{R}^{d_{\theta}}$ to observable quantities. In many scientific and engineering applications, $G$ is defined implicitly through the solution of a system of PDEs. Observations are assumed to be contaminated by additive noise, leading to the \emph{observation model}
\begin{align}
\label{e:data_model}
    y = G(\theta) + \epsilon,
\end{align}
where $\epsilon \sim \mathcal{N}(0, \Sigma_{\epsilon})$ represents Gaussian measurement noise. This induces the likelihood
\begin{align}
    p(y | \theta) = \mathcal{N}\big(y ; G(\theta), \Sigma_{\epsilon}\big).
\end{align}

In addition to observable quantities, we are interested in predicting QoIs $z \in \mathbb{R}^{d_z}$ through a \emph{predictive forward model}
\begin{align}
    z = H(\theta) + \eta,
\label{e:predictive_forward}
\end{align}
where $H$ denotes the deterministic predictive map and $\eta \sim \mathcal{N}(0,\Sigma_{\eta})$ represents predictive or process noise. This formulation induces the conditional predictive distribution
\begin{align}
p(z|\theta)=\mathcal{N}(z; H(\theta), \Sigma_{\eta}).
\end{align}
Although some predictive models may be fully deterministic, such cases lead to degenerate conditional distributions. To ensure that all probability densities and variational objectives remain well-defined, we adopt formulation~\eqref{e:predictive_forward} with a (possibly small) noise term, so that deterministic models are recovered in the vanishing-noise limit.

We further endow the parameters with a prior distribution $p(\theta)$. 
From Bayes’ rule, the posterior distribution of the parameters and the posterior-predictive distribution of the QoIs are given by
\begin{align}
    p(\theta | y) &= \frac{p(y | \theta) p(\theta)}{p(y)}, \label{e:Bayes} \\
    p(z | y) &= \int p(z | \theta) p(\theta | y)\, \mathrm{d}\theta. \label{e:predictive}
\end{align}
The goal of this work is to efficiently approximate both the posterior distribution $p(\theta|y)$ and the posterior-predictive distribution $p(z|y)$.

\subsection{Conventional two-stage variational inference after observing $y$}
\label{ss:classical_variational_method}

Once observational data $y$ becomes available, the conventional approach to Bayesian predictive inference proceeds in two stages. The first stage consists of approximating the posterior distribution $p(\theta | y)$ and generating samples from it. Within a VI framework, this is achieved by introducing a tractable family of distributions $q_{\lambda}(\theta|y)$, parameterized by variational parameters $\lambda$, to approximate the true posterior.

The variational parameters are obtained by minimizing the Kullback--Leibler (KL) divergence between the variational approximation and the true posterior:
\begin{align}
    \lambda^{\ast} 
    &= \argmin_{\lambda} \, \DKL\!\left(q_{\lambda}(\theta|y)\,\|\, p(\theta | y)\right) \nonumber\\
    &= \argmin_{\lambda} \, \mathbb{E}_{q_{\lambda}(\theta|y)} 
    \left[
        \log q_{\lambda}(\theta|y) 
        - \log \left( \frac{p(y | \theta)p(\theta)}{p(y)} \right)
    \right] \nonumber\\
    &= \argmin_{\lambda} \, 
    \mathbb{E}_{q_{\lambda}(\theta|y)} 
    \left[
        \log q_{\lambda}(\theta|y) 
        - \log p(y | \theta) 
        - \log p(\theta)
    \right], 
\label{e:elbo}
\end{align}
which corresponds to minimizing the negative evidence lower bound (ELBO). The marginal likelihood $p(y)$ does not appear in the final optimization since it is independent of $\lambda$. In practice, the expectation in~\eqref{e:elbo} is estimated using Monte Carlo sampling, and the resulting optimization problem is solved using gradient-based methods.

Once the optimal variational approximation $q_{\lambda^{\ast}}(\theta|y)$ is obtained, posterior samples $\{\theta^{(i)}\}_{i=1}^{N_c}$ can be drawn from $q_{\lambda^{\ast}}(\theta|y)$. These samples are then propagated through the predictive model to generate posterior-predictive samples. Specifically, together with samples of the predictive noise $\eta^{(i)}$, evaluations of
\[
z^{(i)} = H(\theta^{(i)}) + \eta^{(i)}
\]
produce samples from the posterior-predictive distribution $p(z|y)$.

This sequential procedure illustrates the separation between posterior inference and predictive uncertainty propagation. The next section develops a variational framework that learns approximations to both distributions simultaneously.

\section{Joint posterior and predictive variational inference}
\label{proposed_vimethod}

\subsection{Upper-bound variational formulation and amortized training}

We first present the variational formulation for a fixed observation $y$, which clarifies the structure of the objective function. We then extend this formulation to an amortized setting that enables offline training prior to observing any data.

Our approach constructs variational approximations to both the posterior and posterior-predictive distributions:
\begin{align}
\label{e:approx_posterior_both}
    p(\theta | y) \approx q_{\lambda}(\theta | y),
    \qquad
    p(z | y) \approx r_{\gamma}(z | y),
\end{align}
where $q_{\lambda}(\theta | y)$ and $r_{\gamma}(z | y)$ are parameterized by variational parameters $\lambda$ and $\gamma$, respectively.

\paragraph{Conditional variational objective}

For a fixed observation $y$, the parameters $\{\lambda,\gamma\}$ can be learned by minimizing discrepancies between the variational distributions and their corresponding targets. In principle this can be achieved by minimizing the KL divergences
\[
\DKL(q_{\lambda}(\theta|y)\,\|\,p(\theta|y)),
\qquad
\DKL(r_{\gamma}(z|y)\,\|\,p(z|y)).
\]

To improve numerical stability and guide learning of the posterior-predictive approximation, we further introduce moment-based regularization. 
The resulting optimization problem can be written as
\begin{align}
\label{e:Gaussian_VI_obj}
\{\lambda^\ast,\gamma^\ast\}
&=
\argmin_{\lambda,\gamma}
\Big[
\DKL(q_{\lambda}(\theta|y)\,\|\,p(\theta|y))
+
\alpha_1 \DKL(r_{\gamma}(z|y)\,\|\,p(z|y))
\nonumber\\ 
& \hspace{5em}+
\text{moment-based regularization}
\Big],
\end{align}
where $\alpha_1 > 0$ is a scalar hyperparameter that controls the relative emphasis on matching the posterior-predictive distribution compared to the posterior.
Specifically, we employ lightweight moment-based regularization that encourages matching of the mean and variance of the posterior-predictive distribution:
\begin{align}
\label{e:reg}
\alpha_2\left\| \mathbb{E}_{r_{\gamma}(z|y)}[z] - \mathbb{E}_{p(z|y)}[z] \right\|_2^2
+
\alpha_3\left\|
\mathbb{E}_{r_{\gamma}(z|y)}
\!\left[
(z-\mathbb{E}_{r_{\gamma}(z|y)}[z])^2
\right]
-
\mathbb{E}_{p(z|y)}
\!\left[
(z-\mathbb{E}_{p(z|y)}[z])^2
\right]
\right\|_2^2,
\end{align}
where $\alpha_2, \alpha_3 \geq 0$ are scalar hyperparameters that control the strength of the moment-matching regularization.
The moments of the variational distribution can often be obtained directly from its parameters, or estimated via Monte Carlo sampling when closed-form expressions are unavailable.
In all experiments, we set $\alpha_1 = \alpha_2 = \alpha_3 = 1$ for simplicity, and do not perform additional hyperparameter tuning.

\paragraph{Upper-bound reformulation}

We now derive tractable expressions for the KL terms. The first KL divergence admits the standard variational form following \eqref{e:elbo}:\begin{align}
\label{e:Gaussian_VI_KL1}
\DKL(q_{\lambda}(\theta|y)\,\|\,p(\theta|y))
=
\mathbb{E}_{q_{\lambda}(\theta|y)}
\!\left[
\log q_{\lambda}(\theta|y)
-
\log p(y|\theta)
-
\log p(\theta)
\right]
+
\log p(y),
\end{align}
where the marginal likelihood $\log p(y)$ is independent of $\lambda$ and $\gamma$ and therefore can be omitted during optimization.

The second KL divergence involves the intractable posterior-predictive distribution:
\begin{align}
\label{e:Gaussian_VI_KL2}
\DKL(r_{\gamma}(z|y)\,\|\,p(z|y))
&=
\mathbb{E}_{r_{\gamma}(z|y)}
\!\left[
\log r_{\gamma}(z|y)
-
\log p(z|y)
\right]
\nonumber\\
&=
\mathbb{E}_{r_{\gamma}(z|y)}
\!\left[
\log r_{\gamma}(z|y)
-
\log \mathbb{E}_{p(\theta|y)}[p(z|\theta)]
\right]
\nonumber\\
&\le
\mathbb{E}_{r_{\gamma}(z|y)}
\!\left[
\log r_{\gamma}(z|y)
-
\mathbb{E}_{p(\theta|y)}[\log p(z|\theta)]
\right]
\nonumber\\
&\approx
\mathbb{E}_{r_{\gamma}(z|y)}
\!\left[
\log r_{\gamma}(z|y)
-
\mathbb{E}_{q_{\lambda}(\theta|y)}[\log p(z|\theta)]
\right].
\end{align}
The inequality follows from Jensen’s inequality, which provides an upper bound that avoids direct evaluation of the intractable quantity 
$\log \mathbb{E}_{p(\theta|y)}[p(z|\theta)]$. 
While this term could in principle be approximated using Monte Carlo together with numerically stabilized log-sum-exp computations \citep{blei2017variational,Blanchard2020}, such estimators typically lead to biased or high-variance gradients when used within stochastic optimization. 
The Jensen bound replaces the logarithm of an expectation with an expectation of log-likelihoods, yielding a tractable and more stable objective for learning the variational posterior-predictive distribution $r_\gamma(z|y)$. 
The bound becomes tight when $p(z|\theta)$ varies weakly over the posterior mass (e.g., when the posterior is concentrated or the predictive model is locally linear). 
In practice, the expectation with respect to the posterior is further approximated using the variational distribution $q_{\lambda}(\theta|y)$.

Combining the above results yields the conditional upper-bound objective
\begin{align}
\label{e:Gaussian_VI_upper_bound}
\mathcal{L}(\lambda,\gamma;y)
&=
\mathbb{E}_{q_{\lambda}(\theta|y)}
\!\left[
\log q_{\lambda}(\theta|y)
-
\log p(y|\theta)
-
\log p(\theta)
\right]
\nonumber\\
&\quad+
\alpha_1
\mathbb{E}_{r_{\gamma}(z|y)}
\!\left[
\log r_{\gamma}(z|y)
-
\mathbb{E}_{q_{\lambda}(\theta|y)}[\log p(z|\theta)]
\right]
\nonumber\\
&\quad+
\text{moment-based regularization}.
\end{align}

\paragraph{Amortized training formulation}

The objective above assumes that $y$ is fixed and known during optimization, which means that the variational problem cannot be set up until the observation becomes available. To enable deployment on yet unseen data, we adopt an amortized formulation in which the variational distributions are trained across possible observations by minimizing the expected loss
\begin{align}
\label{e:Gaussian_VI_upper_bound_amortized}
\{\lambda^\ast,\gamma^\ast\}
=
\argmin_{\lambda,\gamma}
\mathbb{E}_{y}\!\left[
\mathcal{L}(\lambda,\gamma;y)
\right].
\end{align}

This amortized objective can be optimized offline, after which evaluating the posterior and posterior-predictive approximations for a new observation requires only a forward pass through the trained variational distributions. The result is amortized variational approximations $q_{\lambda}(\theta|y)$ and $r_{\gamma}(z|y)$ that can be evaluated for new observation $y$ encountered during deployment.

\subsection{Evaluation of the upper-bound variational objective}
\label{ss:obj_eval}

All terms in the variational objective~\eqref{e:Gaussian_VI_upper_bound_amortized} can be  estimated using Monte Carlo sampling. The outer amortization expectation can be approximated as
\begin{align}
\{\lambda^\ast,\gamma^\ast\}
\approx
\argmin_{\lambda,\gamma}
\frac{1}{N_0}\sum_{i=1}^{N_0}
\mathcal{L}(\lambda,\gamma;y^{(i)}),
\end{align}
where $y^{(i)} \sim p(y)$. These samples can be generated by first drawing $\theta^{(i)} \sim p(\theta)$ from the prior and then propagating them through the observation model~\eqref{e:data_model}. If generating such samples is computationally expensive, domain knowledge or approximations may be used instead to construct representative values of $y$ for amortization.

The loss for each observation can then be estimated as
\begin{align}
\mathcal{L}(\lambda,\gamma;y^{(i)})
&\approx
\frac{1}{N_1} \sum_{j_1=1}^{N_1} 
\left[
\log q_{\lambda}(\theta^{(j_1)}|y^{(i)})
-
\log p(y^{(i)}|\theta^{(j_1)})
-
\log p(\theta^{(j_1)})
\right]
\nonumber\\
&\quad+
\frac{\alpha_1}{N_2} \sum_{j_2=1}^{N_2}
\left[
\log r_{\gamma}(z^{(j_2)}|y^{(i)})
-
\frac{1}{N_3}\sum_{k=1}^{N_3}
\log p(z^{(j_2)}|\theta^{(k)})
\right]
\nonumber\\
&\quad+
\text{moment-based regularization},
\label{e:loss_MC}
\end{align}
where $\theta^{(j_1)}\sim q_{\lambda}(\theta|y^{(i)})$, $z^{(j_2)}\sim r_{\gamma}(z|y^{(i)})$, and $\theta^{(k)}\sim q_{\lambda}(\theta|y^{(i)})$. These samples are straightforward to generate since they only require sampling from the variational distributions and do not involve evaluations of the forward models.
The computational cost of forward model evaluations arises primarily from the likelihood terms $\log p(y^{(i)}|\theta^{(j_1)})$ and $\log p(z^{(j_2)}|\theta^{(k)})$, which require evaluating the forward models $G$ and $H$, respectively.

The moment-based regularization terms can be estimated using standard Monte Carlo estimators
\begin{align}
 \alpha_2\left\| \frac{1}{L_r}\sum_{l_r=1}^{L_r} z^{(l_r)}
-
\frac{1}{L_p}\sum_{l_p=1}^{L_p}z^{(l_p)} \right\|_2^2
+
\alpha_3\left\|
\frac{1}{L_r}\sum_{l_r=1}^{L_r}
\left(z^{(l_r)}-\bar{z}_r\right)^2
-
\frac{1}{L_p}\sum_{l_p=1}^{L_p}
\left(z^{(l_p)}-\bar{z}_p\right)^2
\right\|_2^2,
\label{e:reg_MC}
\end{align}
where $z^{(l_r)}\sim r_{\gamma}(z|y^{(i)})$ and $z^{(l_p)}\sim p(z|y^{(i)})$, and $\bar{z}_r$ and $\bar{z}_p$ denote the corresponding Monte Carlo empirical means. Since samples from $r_{\gamma}$ are inexpensive to generate, $L_r$ can be chosen large. In contrast, generating $z^{(l_p)}$ requires evaluating the predictive model $H$ and may therefore be computationally expensive, so $L_p$ is typically chosen to be relatively small.

\paragraph{Special case: Independent Gaussian models}

For common choices of variational families and prior distributions, several terms in the objective admit closed-form expressions. We summarize these computations for independent Gaussian variational approximations:
\begin{align}
q_{\lambda}(\theta|y)
&=
\mathcal{N}\!\left(
\theta;\mu_{\theta}(\lambda;y),
\operatorname{diag}(\sigma_{\theta}^2(\lambda;y))
\right), \label{e:q_Gaussian}
\\
r_{\gamma}(z|y)
&=
\mathcal{N}\!\left(
z;\mu_{z}(\gamma;y),
\operatorname{diag}(\sigma_{z}^2(\gamma;y))
\right).\label{e:r_Gaussian}
\end{align}
For these distributions the entropy terms admit closed-form expressions
\begin{align}
\mathbb{E}_{q_{\lambda}}[\log q_{\lambda}]
&=
-\frac{d_\theta}{2}(1+\log 2\pi)
-\frac{1}{2}\sum_{i=1}^{d_\theta}\log\sigma_{\theta,i}^2,
\\
\mathbb{E}_{r_{\gamma}}[\log r_{\gamma}]
&=
-\frac{d_z}{2}(1+\log 2\pi)
-\frac{1}{2}\sum_{i=1}^{d_z}\log\sigma_{z,i}^2.
\end{align}
If the prior is Gaussian $p(\theta)=\mathcal{N}(\theta;\mu_0,\Sigma_0)$, the prior expectation becomes
\begin{align}
\mathbb{E}_{q_{\lambda}}[\log p(\theta)]
&=
-\frac{d_\theta}{2}\log(2\pi)
-
\frac{1}{2}\log\det(\Sigma_0)
\nonumber\\
&\quad-
\frac{1}{2}\Big(
\operatorname{tr}(\Sigma_0^{-1}\operatorname{diag}(\sigma_\theta^2))
+
(\mu_\theta-\mu_0)^\top\Sigma_0^{-1}(\mu_\theta-\mu_0)
\Big).
\end{align}

\paragraph{Special case: Log-normal posterior-predictive distributions}

For predictive quantities known to be strictly positive, we alternatively consider a log-normal variational model
\begin{align}
r_{\gamma}(z|y)
=
\operatorname{LogNormal}
\!\left(
z;\mu_z(y;\gamma),
\operatorname{diag}(\sigma_z^2(y;\gamma))
\right),\label{e:r_logNormal}
\end{align}
where $\mu_z$ and $\sigma_z$ correspond to the mean and standard deviation of the underlying Gaussian distribution in log-space.
In this case the only modification in the variational objective is the entropy term
\begin{align}
\mathbb{E}_{r_{\gamma}}[\log r_{\gamma}]
=
-\sum_{i=1}^{d_z}
\left(
\mu_{z,i}
+
\frac{1}{2}
+
\frac{1}{2}\log(2\pi\sigma_{z,i}^2)
\right),
\end{align}
while all remaining expectations are computed using the same Monte Carlo estimators as in the Gaussian case.

\subsection{Optimization and training}

The variational parameters $\{\lambda,\gamma\}$ are optimized using stochastic gradient-based methods. We employ the Adam optimizer~\citep{kingma2015adam} together with mini-batch stochastic gradients. Gradients are computed using automatic differentiation in TensorFlow~\citep{tensorflow2015-whitepaper}. Expectations appearing in the objective are evaluated either analytically when possible, or approximated using Monte Carlo sampling, as described above.

To enable gradient-based optimization through stochastic sampling, we employ the standard reparameterization trick. For example,
\begin{align}
\theta
=
\mu_\theta(\lambda;y)
+
\sigma_\theta(\lambda;y)\odot\epsilon,
\qquad
\epsilon\sim\mathcal{N}(0,I),
\end{align}
which allows gradients to propagate through the sampling operation.

Network parameters are initialized using He initialization~\citep{He2015Delving}. Training proceeds for a fixed number of epochs until convergence.

\begin{algorithm}
\caption{Amortized joint posterior and predictive VI.}
\label{alg:vbi1}
\begin{algorithmic}[1]
\State Initialize variational parameters $\lambda,\gamma$ and Monte Carlo sample sizes.
\State Draw amortization observations $\{y^{(i)}\}_{i=1}^{N_0}\sim p(y)$.
\For{$l=1,\ldots,L$}
    \State Shuffle observations and form mini-batches.
    \For{each mini-batch}
        \State Estimate stochastic gradients of the loss~\eqref{e:loss_MC} via Monte Carlo sampling and automatic differentiation.
        \State Update $(\lambda,\gamma)$ using Adam.
    \EndFor
\EndFor
\State Return trained amortized distributions $q_{\lambda}(\theta|y)$ and $r_{\gamma}(z|y)$.
\end{algorithmic}
\end{algorithm}

\section{Numerical examples}
\label{numerical_examples}

We evaluate the performance of the proposed method through a sequence of numerical examples of increasing complexity, ranging from analytical benchmarks to high-fidelity engineering models. 
Case~1 considers two one-dimensional (1D) benchmarks. Case~1a is a linear-Gaussian problem for which both the posterior and posterior-predictive distributions admit closed-form expressions, enabling direct verification of the method. Case~1b introduces nonlinear observation and prediction models, leading to non-Gaussian posterior and posterior-predictive distributions. 
Case~2 extends the problem to a two-dimensional (2D) parameter and observation space with nonlinear models. 
Case~3 further examines scalability in higher-dimensional linear settings, with dimensions up to 20. 
Finally, Case~4 considers a finite-element solid mechanics problem governed by a PDE, representing a realistic high-fidelity engineering application.

For each case, we compare the posterior-predictive distributions produced by the proposed method with those obtained from the conventional two-stage VI procedure described in Section~\ref{ss:classical_variational_method}. 
Both approaches are further compared against reference solutions obtained either analytically when available or otherwise using high-quality Monte Carlo estimates based on MCMC posterior samples.

To quantify the accuracy of the posterior-predictive distributions, we estimate the KL divergence between the approximate and reference distributions using Monte Carlo sampling. 
Additionally, we report relative errors in the predictive mean and variance to assess moment-level accuracy.

\paragraph{Sampling configuration}

Sampling configuration. Unless otherwise stated, the following sampling settings are used throughout the numerical experiments. For the proposed method, amortized training uses $N_0=10^5$ samples of $y$, while the internal Monte Carlo estimators use $N_1=N_2=L_r=10^4$ and $N_3=L_p=10^3$. For the conventional two-stage VI approach, $N_c=10^5$ posterior samples are used for predictive propagation. Reference solutions are computed using $10^5$ MCMC posterior samples.

Table~\ref{table:GH_evaluations} summarizes the number of $G$ and $H$ evaluations required by the conventional and proposed methods, separated into offline and online phases. These counts should be interpreted as an offline--online cost profile rather than a direct measure of total computational efficiency. The proposed method learns amortized predictive distributions over possible observations and therefore incurs additional offline evaluations of $H$. In contrast, the conventional two-stage method performs inference only after a specific observation $y$ is available and requires predictive model evaluations during online inference. Thus, the main computational benefit of the proposed method is that, after offline training, posterior and posterior-predictive inference for a new observation can be obtained without additional evaluations of $G$ or $H$.

\begin{table}[htbp]
\caption{Number of $G$ and $H$ function evaluations per optimization iteration for the conventional and proposed methods, separated into offline and online phases. Both methods are run for 3{,}200 optimization iterations. The table illustrates the offline--online cost allocation rather than a direct total-cost comparison.}
\centering
\begin{tabular}{|c|c|c|c|c|} 
 \hline
   & \multicolumn{2}{c|}{\textbf{Conventional method}} & \multicolumn{2}{c|}{\textbf{Proposed method}} \\ 
 \hline
    & $G$ & $H$ & $G$ & $H$ \\ 
 \hline
 Offline & $6.4\times10^5$ & $0$ & $6.4\times10^5$ & $6.4\times10^{8}$ \\
\hline
 Online & $0$ & $10^5$ for Cases 1--2; $10^4$ for Case 3--4 & $0$ & $0$ \\
\hline
\end{tabular}
\label{table:GH_evaluations}
\end{table}

\paragraph{Variational distribution architecture}

The neural network architectures used to parameterize the variational distributions are summarized in Table~\ref{table:hyperparameter}. 
Independent Gaussian variational distributions are employed in Cases~1--3. For Case~4, which involves strictly positive physical quantities, a log-normal predictive distribution is adopted.
In all cases, the neural networks output the corresponding distribution parameters (mean and standard deviation), following \eqref{e:q_Gaussian}, \eqref{e:r_Gaussian}, and \eqref{e:r_logNormal}.

All experiments are implemented in Python using the TensorFlow framework~\citep{tensorflow2015-whitepaper}. 
Computations are performed on the University of Michigan Great Lakes High Performance Computing Cluster and Google Cloud Compute Engine nodes equipped with NVIDIA A40 or Tesla V100 GPUs.

\begin{table}[htbp]
\caption{Neural network hyperparameters used to parameterize the variational distributions in the numerical experiments.}
\centering
\begin{tabular}{|c|c|c|} 
 \hline
   & \textbf{Case 1a, 3} & \textbf{Cases 1b, 2, 4} 
   \\ 
 \hline
 \# of hidden layers & 1 & 3 
 \\
\hline
 \# of neurons per hidden layer & \multicolumn{2}{c|}{20} 
 \\
\hline
Activation & \multicolumn{2}{c|}{ReLU} 
\\
 \hline
\end{tabular}
\label{table:hyperparameter}
\end{table}

\subsection{Case 1: 1D linear and nonlinear benchmarks}

We first consider two 1D benchmark problems to evaluate the proposed method in both linear-Gaussian and nonlinear settings. 
The linear case admits closed-form posterior and posterior-predictive distributions and therefore provides a baseline for verifying the correctness of the proposed approach. 
The nonlinear case then assesses performance when analytical solutions are no longer available.

\paragraph{Case 1a: Linear benchmark}

We begin with a linear-Gaussian problem defined by the observation and predictive models
\begin{align}
\label{e:case1_models}
y = 2\theta + \epsilon, \qquad z = 3\theta + \eta,
\end{align}
where $\epsilon \sim \mathcal{N}(0,\sigma_\epsilon^2=10^{-4})$ and $\eta \sim \mathcal{N}(0,\sigma_\eta^2=10^{-3})$ denote observation and predictive noise, respectively. 
Assuming a standard Gaussian prior $\theta\sim \mathcal{N}(0,1)$, the posterior distribution can be obtained analytically as
\begin{align}
p(\theta|y)
=
\mathcal{N}
\left(
\theta; 
\frac{2y}{4+\sigma_{\epsilon}^2},
\,
(1+4\sigma_\epsilon^{-2})^{-1}
\right),
\end{align}
and the corresponding posterior-predictive distribution is
\begin{align}
p(z|y)
=
\mathcal{N}
\left(
z;
\frac{6y}{4+\sigma_{\epsilon}^2},
\,
9(1+4\sigma_\epsilon^{-2})^{-1}+\sigma_\eta^{2}
\right).
\end{align}
These closed-form expressions provide reference solutions for evaluating the accuracy of the conventional and proposed methods.

\begin{figure}[htbp]
\centering
\includegraphics[scale=0.5]{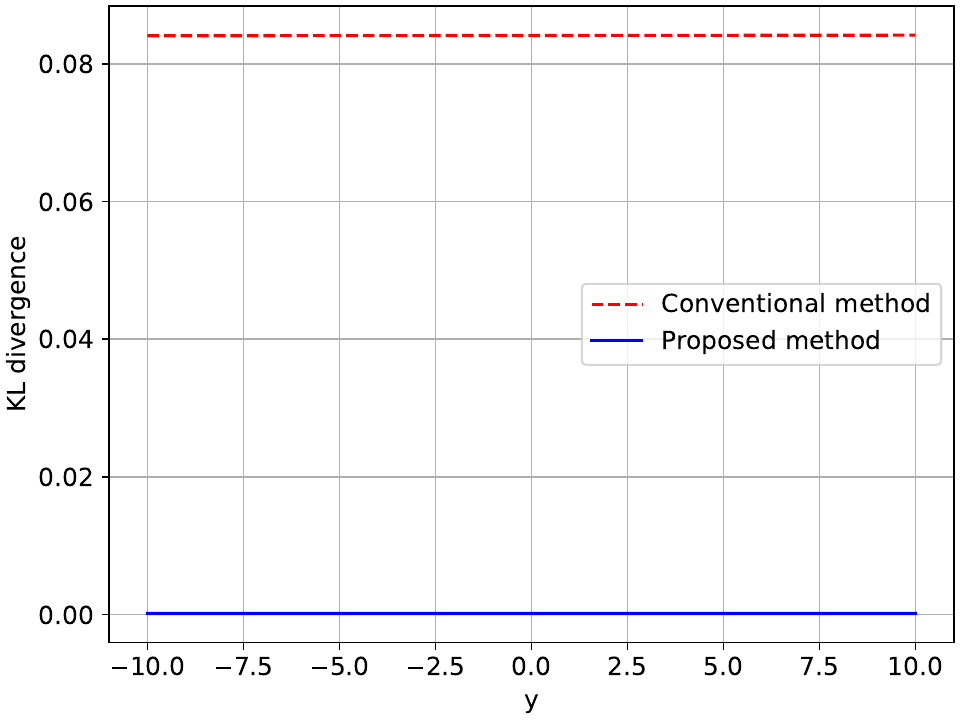}
\caption{Case 1a: KL divergence between the approximate and reference posterior-predictive distributions for the conventional and proposed methods.}
\label{fig:kld_case1}
\end{figure}

\begin{figure}[htbp]
\centering
\includegraphics[scale=0.48]{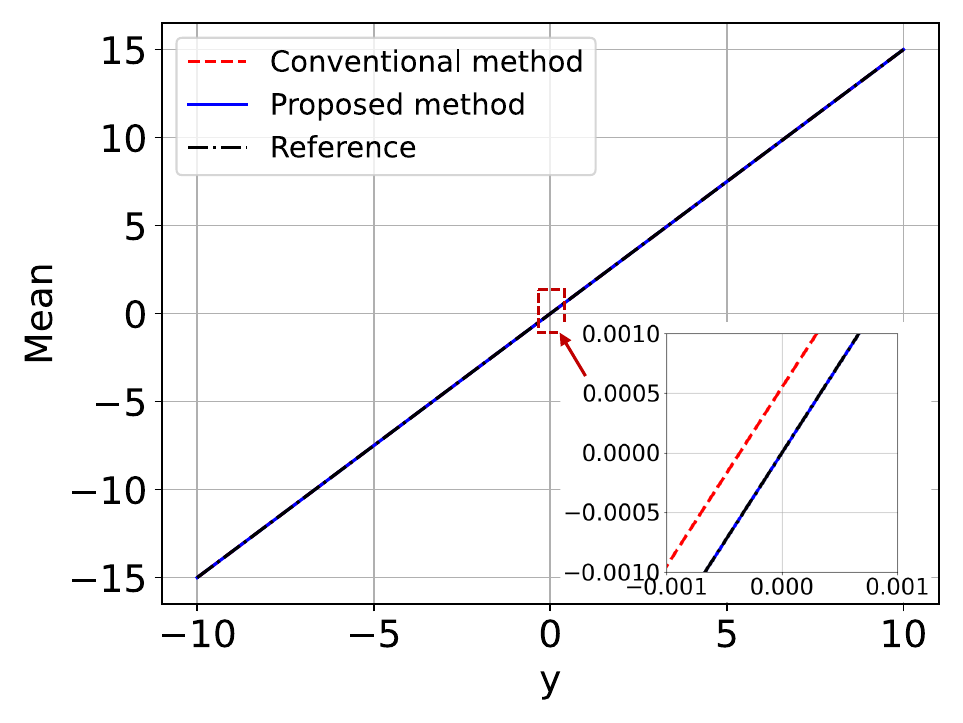}
\includegraphics[scale=0.48]{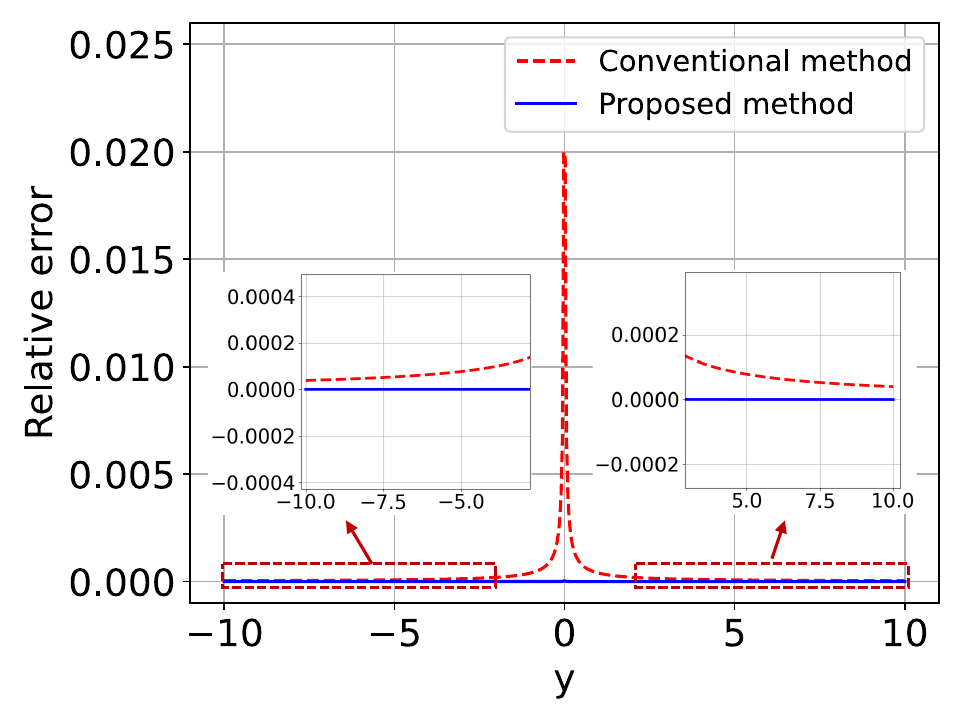}
\caption{Case 1a: (left) Mean of the approximate posterior-predictive distributions; (right) corresponding relative errors for the conventional and proposed methods.}
\label{fig:mean_linear_case1}
\end{figure}

\begin{figure}[htbp]
\centering
\includegraphics[scale=0.48]{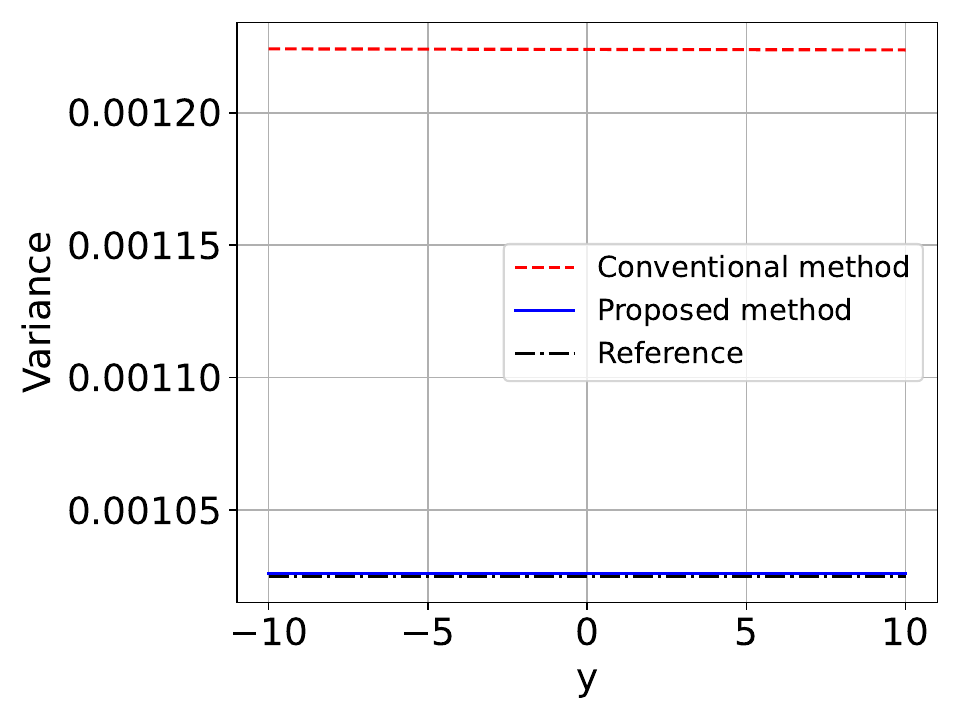}
\includegraphics[scale=0.48]{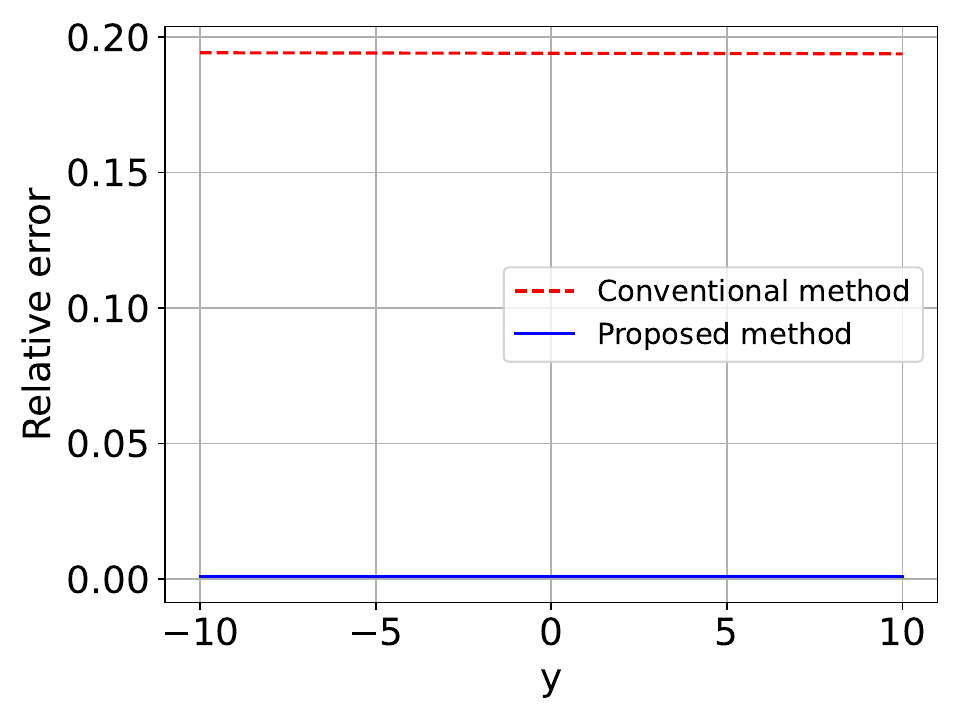}
\caption{Case 1a: (left) Variance of the approximate posterior-predictive distributions; (right) corresponding relative errors for the conventional and proposed methods.}
\label{fig:sig_linear_case1}
\end{figure}

Figure~\ref{fig:kld_case1} reports the KL divergence between the approximate posterior-predictive distributions and the analytical reference solution. 
The proposed method yields consistently smaller and near-zero KL divergence values across the tested values of $y$, indicating a highly accurate predictive approximation.
Figures~\ref{fig:mean_linear_case1} and~\ref{fig:sig_linear_case1} further compare the predicted mean and variance of the posterior-predictive distribution with the analytical reference values. 
The proposed method exhibits closer agreement with the reference solutions and achieves smaller relative errors than the conventional approach.

In addition, due to the amortized formulation, the proposed method requires only a single offline training stage. 
After training, predictive inference can be obtained by directly evaluating the learned variational distributions at the observed $y$. 
In contrast, the conventional two-stage approach must conduct the inference only after $y$ is observed.

\paragraph{Case 1b: Nonlinear benchmark}

Next, we consider a nonlinear and non-Gaussian problem defined by
\begin{align}
y = 0.2\theta^2 + 0.1 + \epsilon, \qquad
z = e^{\theta} + 0.2 + \eta,
\end{align}
where $\epsilon \sim \mathcal{N}(0,\sigma_{\epsilon}^2=10^{-2})$ and $\eta \sim \mathcal{N}(0,\sigma_{\eta}^2=10^{-3})$.

\begin{figure}[htbp]
\centering
\includegraphics[scale=0.5]{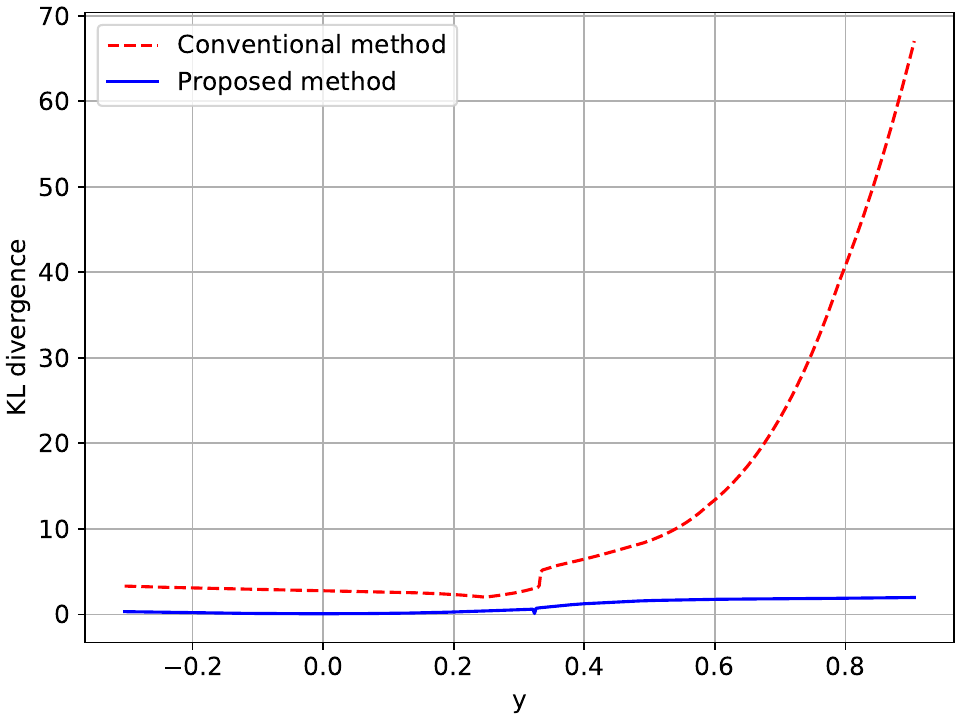}
\caption{Case 1b: KL divergence between the approximate and reference posterior-predictive distributions for the conventional and proposed methods.}
\label{fig:kld_case2}
\end{figure}

\begin{figure}[htbp]
\centering
\includegraphics[scale=0.48]{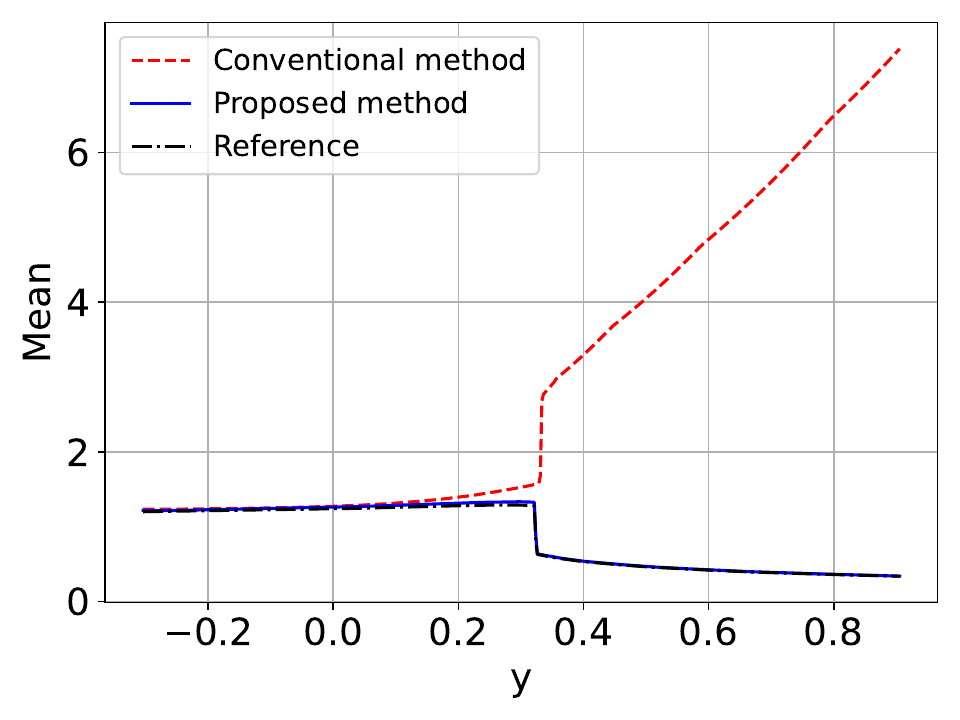}
\includegraphics[scale=0.48]{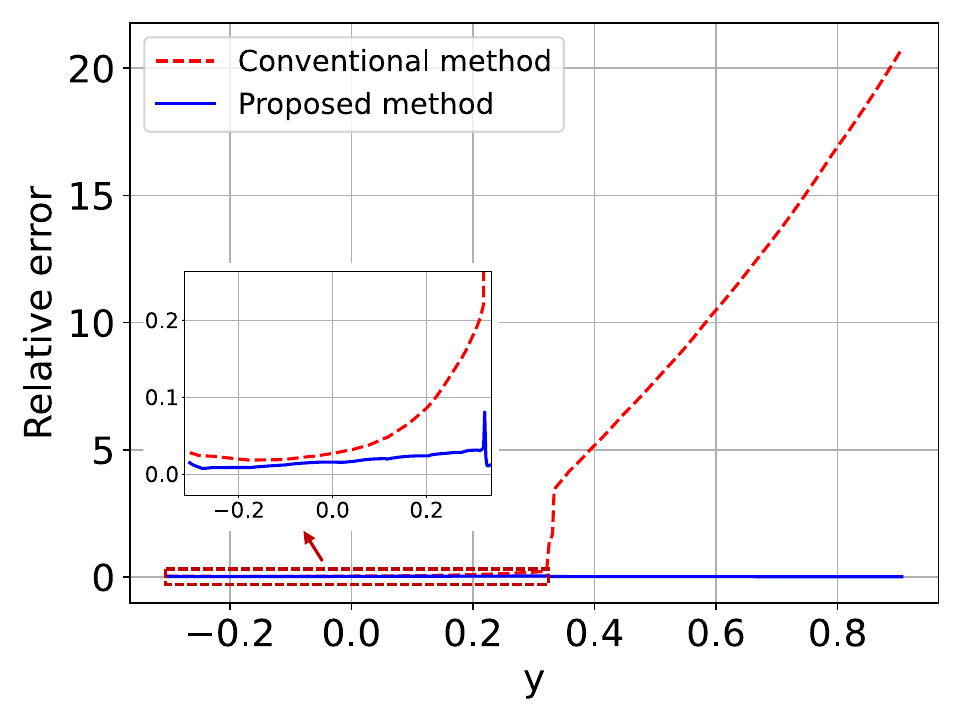}
\caption{Case 1b: (left) Mean of the approximate posterior-predictive distributions; (right) corresponding relative errors for the conventional and proposed methods.}
\label{fig:mean_nonlinear_case2}
\end{figure}

\begin{figure}[htbp]
\centering
\includegraphics[scale=0.48]{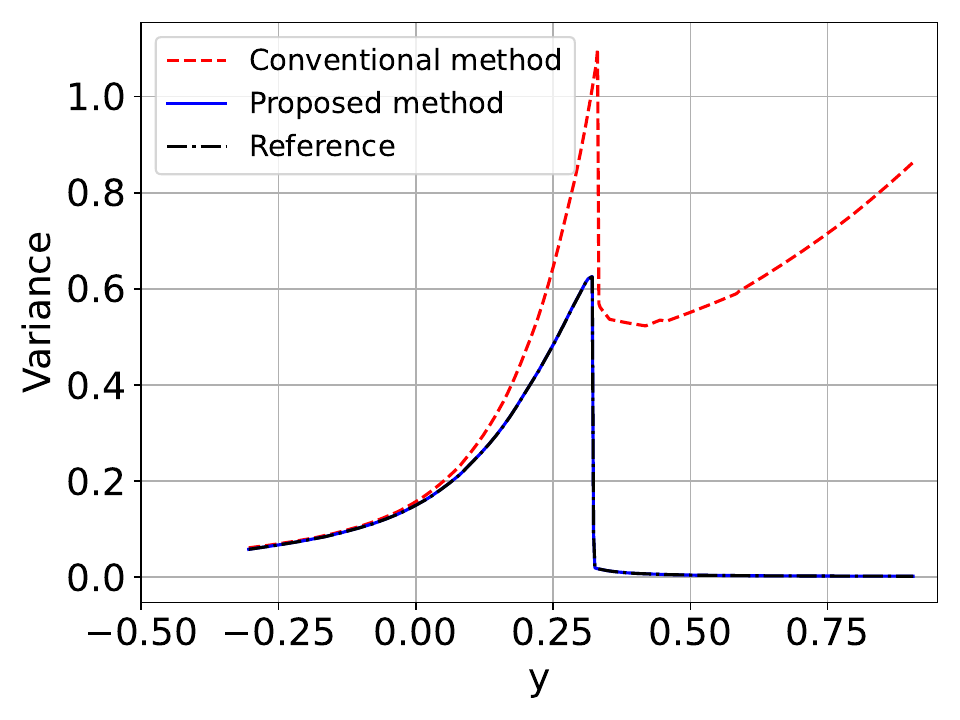}
\includegraphics[scale=0.48]{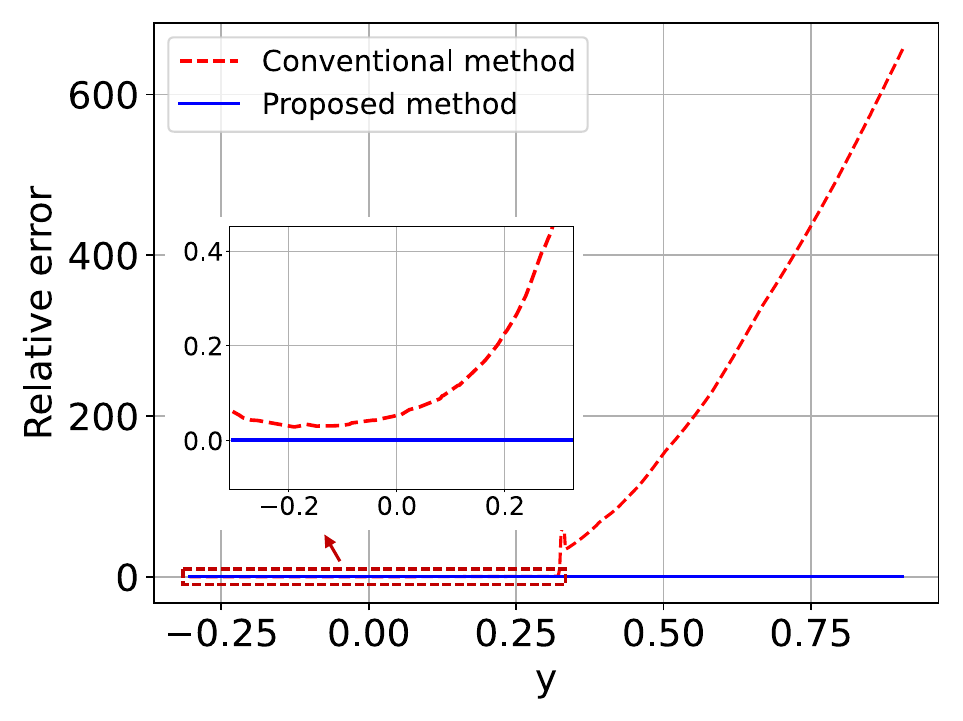}
\caption{Case 1b: (left) Variance of the approximate posterior-predictive distributions; (right) corresponding relative errors for the conventional and proposed methods.}
\label{fig:sig_nonlinear_case2}
\end{figure}

Figure~\ref{fig:kld_case2} shows the KL divergence between the approximate and reference predictive distributions across different observations $y$. 
The proposed method again yields substantially smaller KL divergence values than the conventional approach.
Figures~\ref{fig:mean_nonlinear_case2} and~\ref{fig:sig_nonlinear_case2} compare the predicted mean and variance of the posterior-predictive distributions with the reference values. 
The proposed method demonstrates improved agreement with the reference solutions and consistently smaller relative errors than the conventional method.

\subsection{Case 2: 2D nonlinear and non-Gaussian problem}

We next consider a 2D nonlinear problem to evaluate the method in a higher-dimensional setting.
The observation and predictive models are defined as
\begin{align}
\begin{bmatrix}
y_1 \\
y_2
\end{bmatrix}
=
\begin{bmatrix}
2\theta_1^2+\theta_2+2 \\
\theta_2+\theta_2^4+\theta_1+1
\end{bmatrix}
+
\begin{bmatrix}
\epsilon_1 \\
\epsilon_2
\end{bmatrix},
\qquad
\begin{bmatrix}
z_1 \\
z_2
\end{bmatrix}
=
\begin{bmatrix}
e^{\theta_1}+\theta_2+0.2 \\
e^{\theta_2}+\theta_1+0.1
\end{bmatrix}
+
\begin{bmatrix}
\eta_1 \\
\eta_2
\end{bmatrix},
\end{align}
where $\epsilon_1,\epsilon_2 \sim \mathcal{N}(0,\sigma_{\epsilon}^2=10^{-1})$ and $\eta_1,\eta_2 \sim \mathcal{N}(0,\sigma_{\eta}^2=10^{-2})$.

\begin{figure}[htbp]
\centering
\includegraphics[scale=0.48]{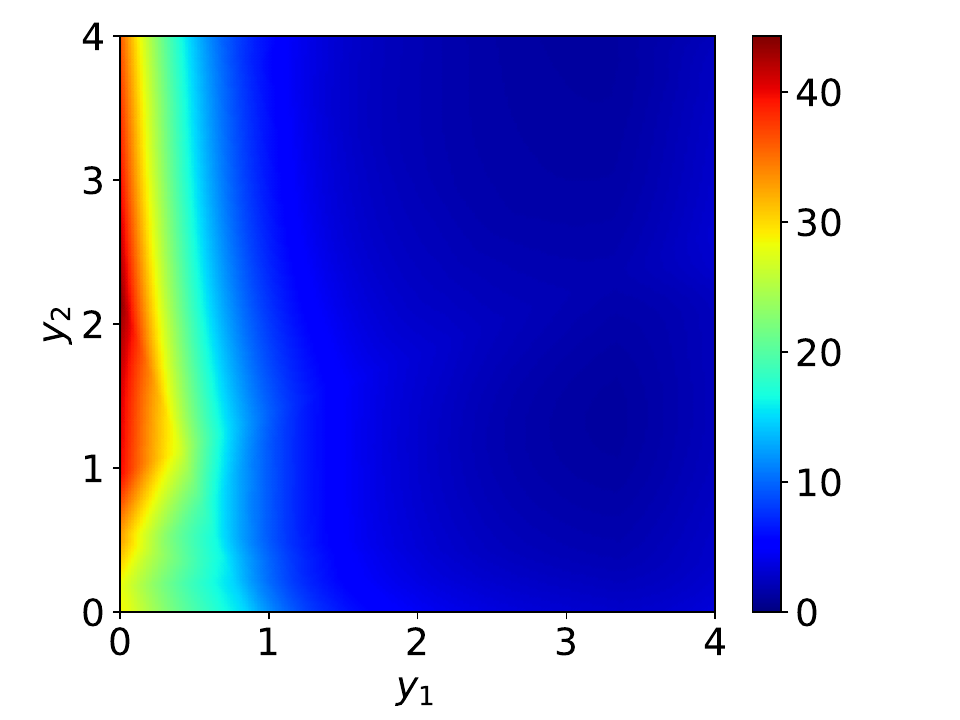}
\includegraphics[scale=0.48]{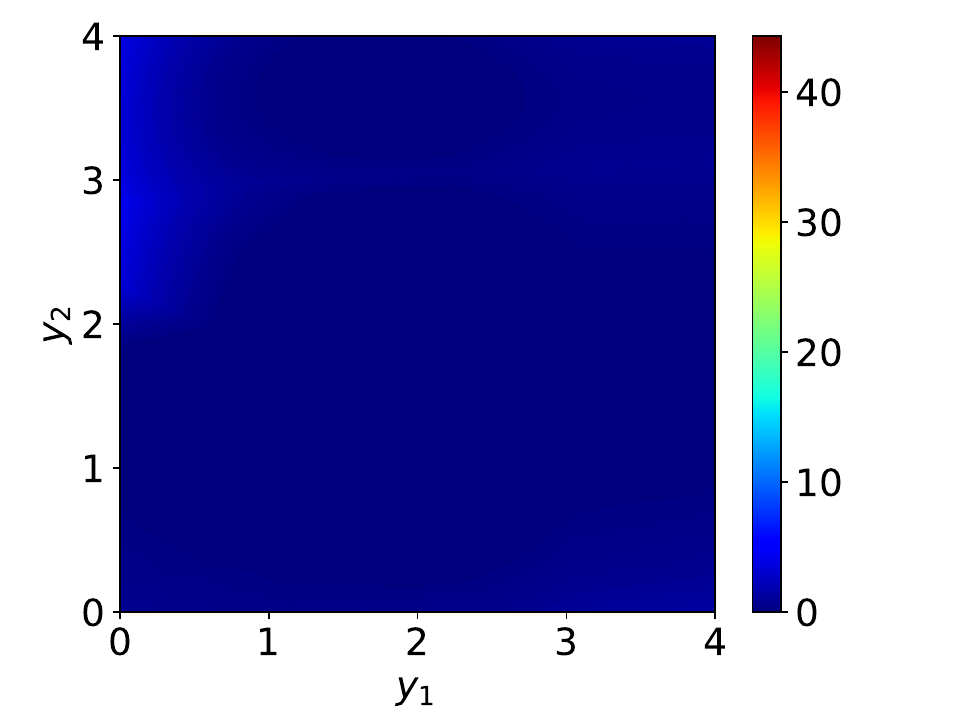}
\caption{Case 2: KL divergence between the approximate and reference posterior-predictive distributions for the (left) conventional and (right) proposed methods.}
\label{fig:kld_case3}
\end{figure}

Figure~\ref{fig:kld_case3} shows the KL divergence between the approximate and reference predictive distributions. 
The proposed method consistently produces smaller KL divergence values across different observations $y$. 
In particular, the maximum KL divergence for the conventional method exceeds 40, whereas the proposed method maintains values below 5.

\begin{figure}[htbp]
\includegraphics[scale=0.6]{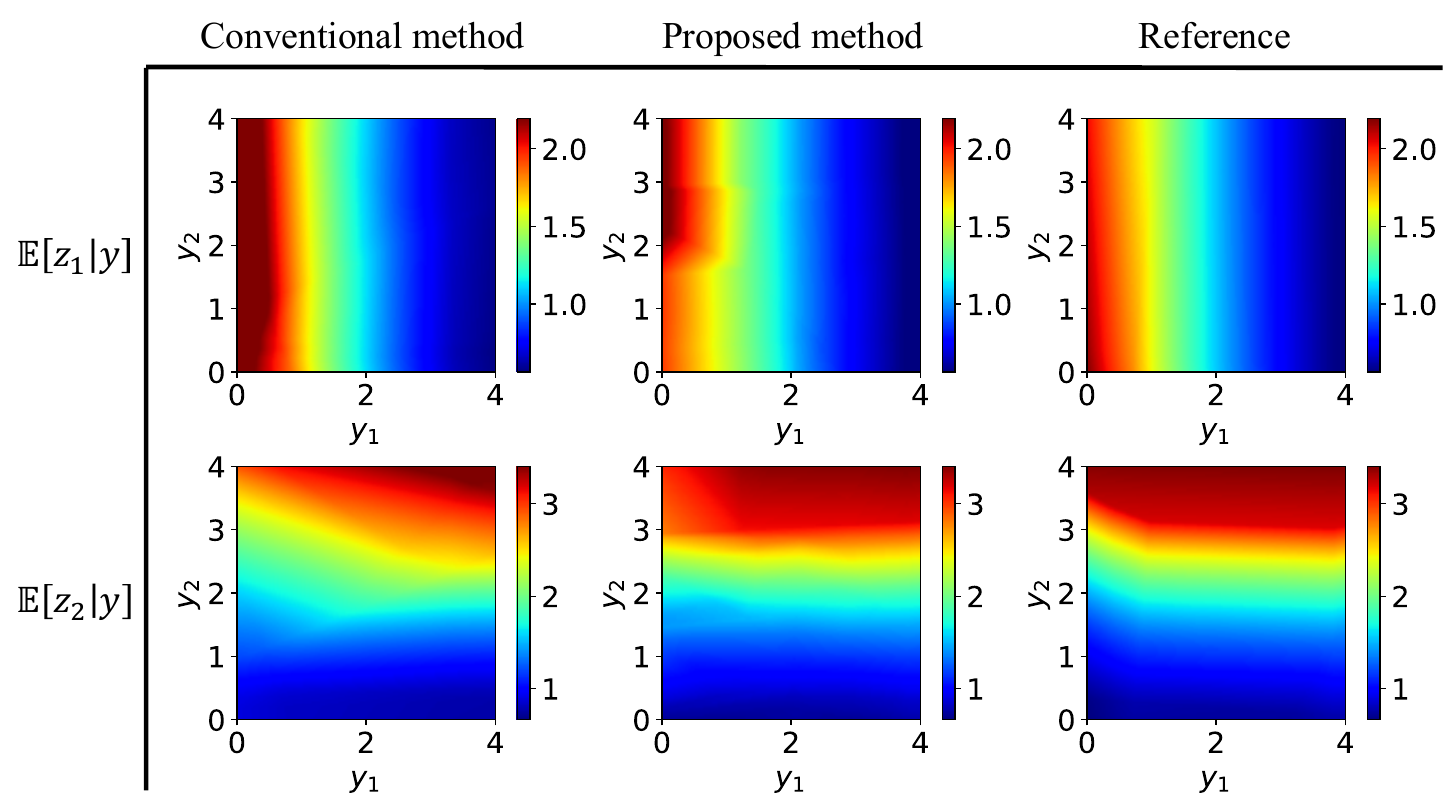}
\centering
\caption{Case 2: Mean of the posterior-predictive distributions. Rows correspond to the two components $z_1$ and $z_2$, and columns show the conventional, proposed, and reference solutions.}
\label{fig:mean_case3}
\end{figure}

\begin{figure}[htbp]
\includegraphics[scale=0.6]{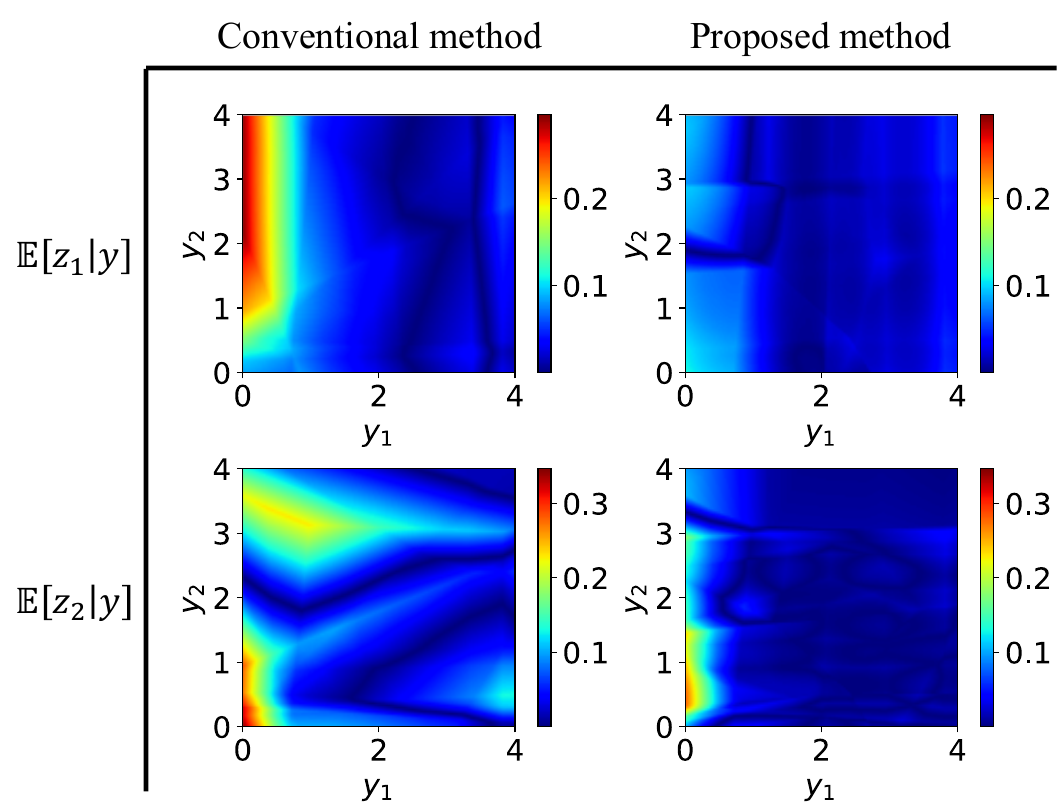}
\centering
\caption{Case 2: Relative errors of the posterior-predictive mean. Rows correspond to the two components $z_1$ and $z_2$, and columns show the conventional and proposed  methods.}
\label{fig:rela_mean_case3}
\end{figure}

\begin{figure}[htbp]
\includegraphics[scale=0.6]{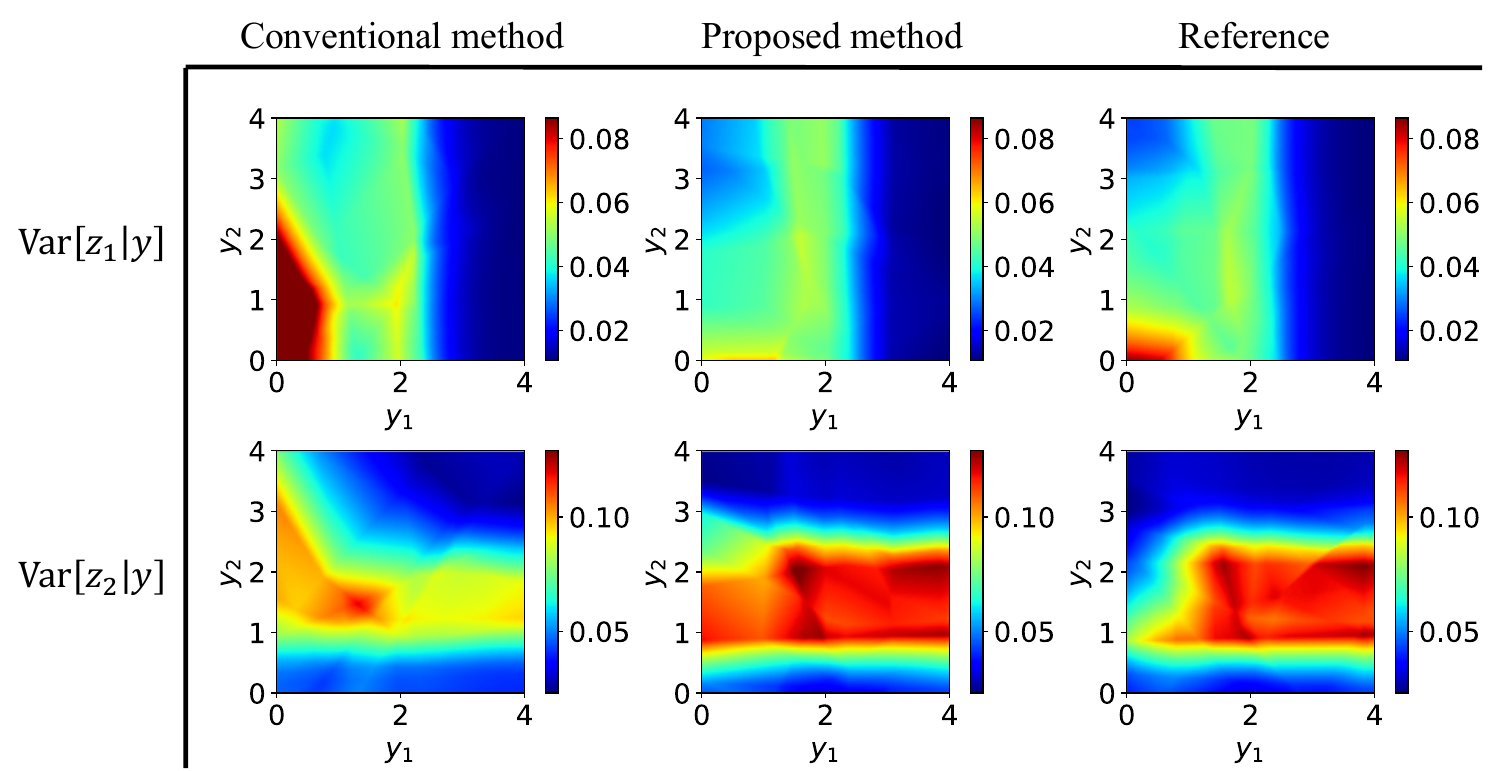}
\centering
\caption{Case 2: Variance of the posterior-predictive distributions. Rows correspond to the two components $z_1$ and $z_2$, and columns show the conventional, proposed, and reference solutions.}
\label{fig:sig_case3}
\end{figure}

\begin{figure}[htbp]
\includegraphics[scale=0.6]{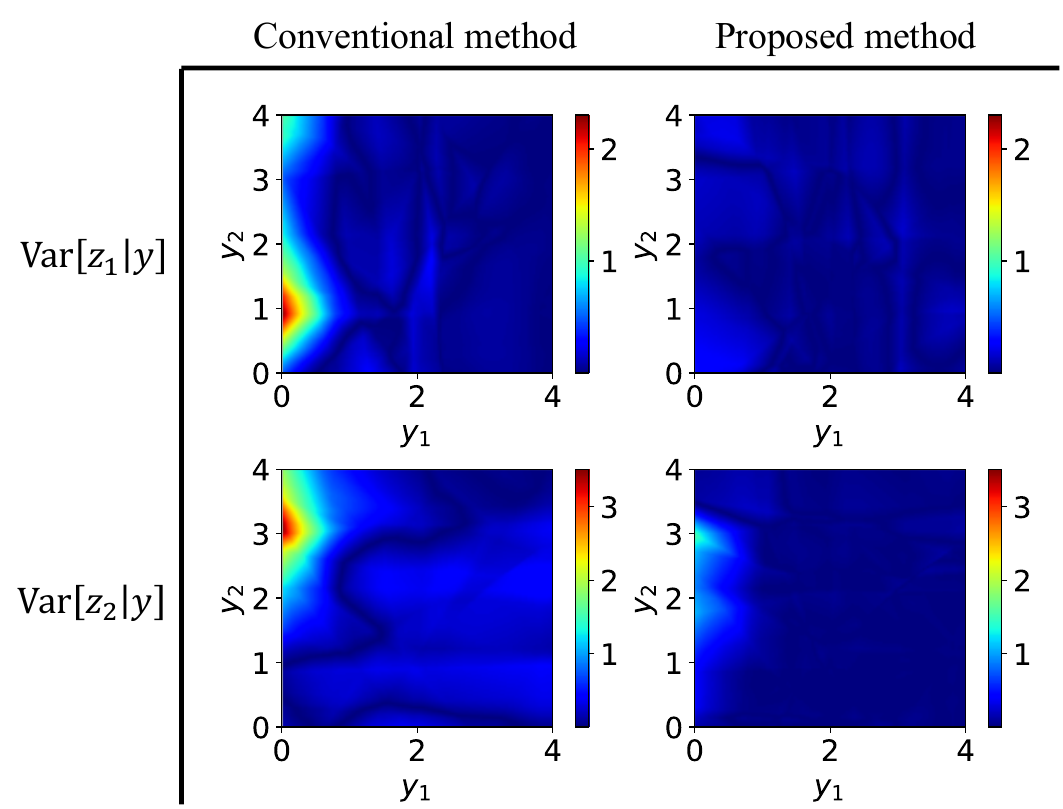}
\centering
\caption{Case 2: Relative errors of the posterior-predictive variance. Rows correspond to the two components $z_1$ and $z_2$, and columns show the conventional and proposed methods.}
\label{fig:rela_sig_case3}
\end{figure}

Figures~\ref{fig:mean_case3} and~\ref{fig:rela_mean_case3} compare the predicted means and their relative errors. 
The proposed method achieves significantly better agreement with the reference solution, with maximum relative errors below $10\%$, compared with over $30\%$ for the conventional approach.
Similarly, Figures~\ref{fig:sig_case3} and~\ref{fig:rela_sig_case3} present the predicted variances and their corresponding relative errors. 
The proposed method again demonstrates improved accuracy, maintaining relative errors below $10\%$, whereas the conventional variational method exhibits errors exceeding $100\%$.

\begin{figure}[htbp]
\includegraphics[scale=0.65]{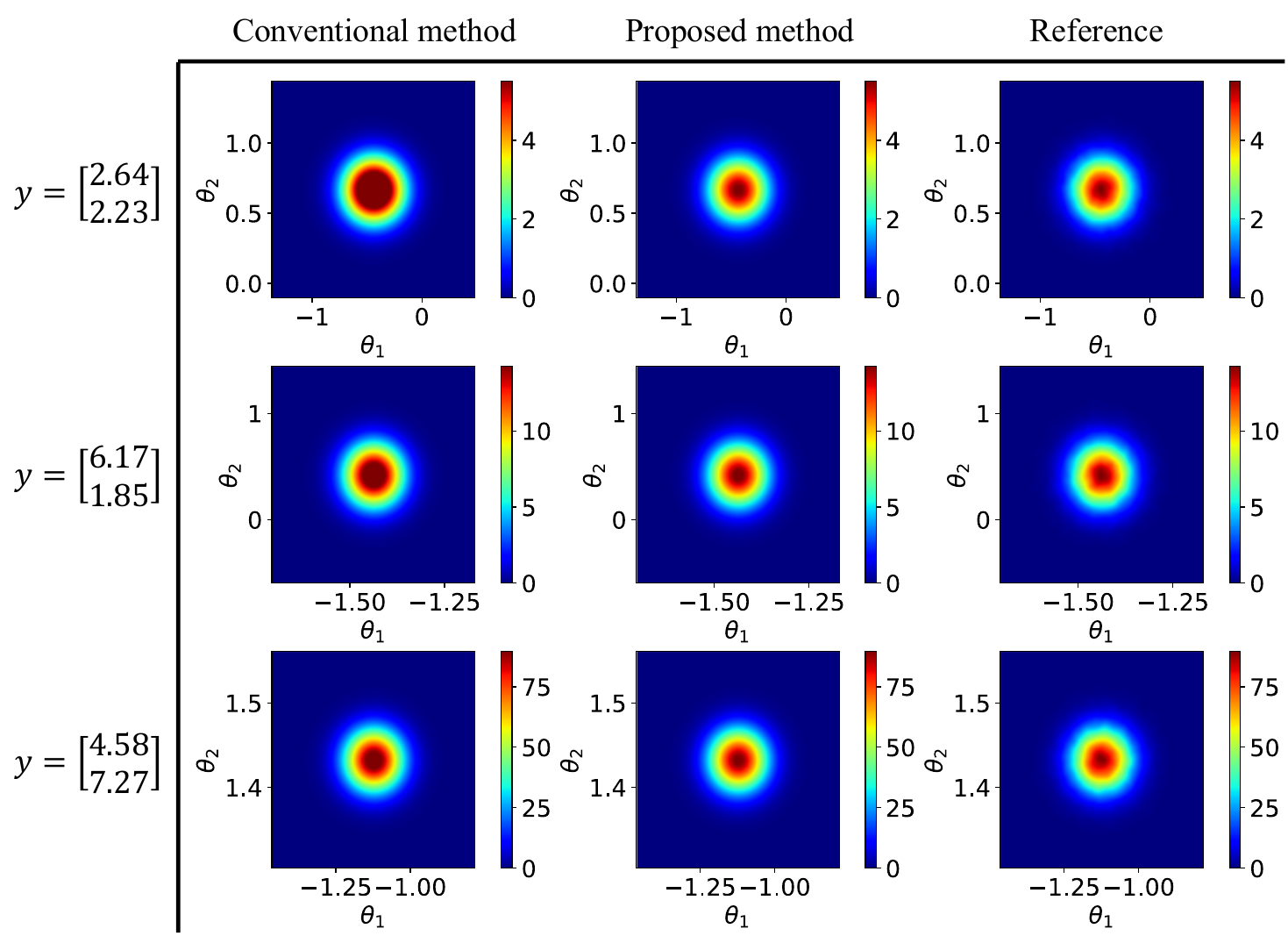}
\centering
\caption{Case 2: Examples of posterior distributions for different values of $y$. Rows correspond to the two components $\theta_1$ and $\theta_2$, and columns show the conventional, proposed, and reference methods.}
\label{fig:p_theta_y_case3}
\end{figure}

\begin{figure}[htbp]
\includegraphics[scale=0.65]{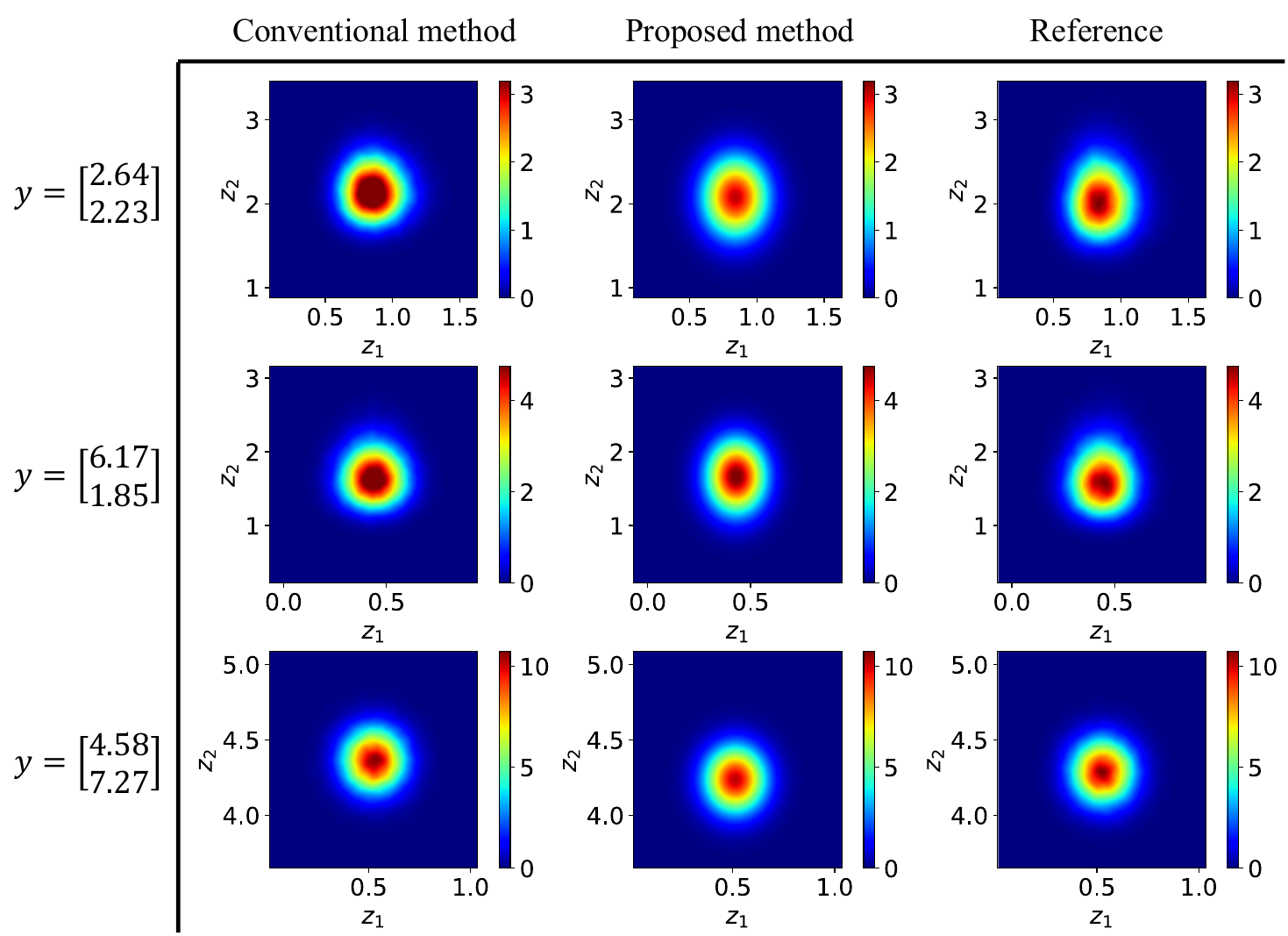}
\centering
\caption{Case 2: Examples of posterior-predictive distributions for different values of $y$. Rows correspond to the two components $z_1$ and $z_2$, and columns show the conventional, proposed, and reference methods.}
\label{fig:p_z_y_case3}
\end{figure}

Figure~\ref{fig:p_theta_y_case3} and~\ref{fig:p_z_y_case3} illustrate representative posterior and posterior-predictive distributions for different values of $y$. Overall, the proposed method achieves closer agreement with the reference solutions than the conventional method, particularly in regions of high probability density.

\subsection{Case 3: High-dimensional linear problems}

We extend the proposed method to higher-dimensional linear problems with $d_{\theta} \in \{5, 10, 15, 20\}$. The observation and predictive models are defined as
\begin{align}
y = A\theta + \epsilon, \qquad
z = B\theta + \eta,
\end{align}
where $\epsilon \sim \mathcal{N}(0,\sigma_{\epsilon}^2=10^{-4})$ and $\eta \sim \mathcal{N}(0,\sigma_{\eta}^2=10^{-3})$. The matrices $A, B \in \mathbb{R}^{d \times d}$ have entries independently sampled from the uniform distribution on $[0,2]$.

For this linear-Gaussian setting, both the posterior and posterior-predictive distributions admit closed-form expressions. The posterior distribution is given by
\begin{align}
p(\theta|y)
=
\mathcal{N}
\left(
\theta; 
\sigma_{\epsilon}^{-2}\Sigma_1 A^\top y,
\,
\Sigma_1
\right),
\qquad
\Sigma_1 =
\left(
I + \sigma_{\epsilon}^{-2} A^\top A
\right)^{-1},
\end{align}
and the corresponding posterior-predictive distribution is
\begin{align}
p(z|y)
=
\mathcal{N}
\left(
z; 
\sigma_{\epsilon}^{-2} B \Sigma_1 A^\top y,
\,
\sigma_{\eta}^2 I + B \Sigma_1 B^\top
\right).
\end{align}
These analytical expressions provide reference solutions for evaluating the accuracy of the conventional and proposed methods. 

Table~\ref{table:error_case4} reports the average KL divergence and relative errors of the posterior-predictive mean and variance across different observations $y$. 
Across all problem dimensions, the proposed method consistently achieves smaller KL divergence and lower relative errors compared to the conventional method, demonstrating its ability to maintain accuracy as the problem dimension increases in linear settings.

\begin{table}[htbp]
\caption{Case 3: Average KL divergence and relative errors of the posterior-predictive mean and variance for the conventional and proposed methods across measurement data in $d_{\theta}\in \{5, 10, 15, 20\}$-dimensional settings.}
\centering
\begin{tabular}{|C{2em}|C{4cm}|C{3cm}|C{3cm}|C{3cm}|} 
 \hline
   $d_{\theta}$ & \textbf{Method} & \textbf{KL divergence} & \textbf{Relative error of mean} & \textbf{Relative error of variance}\\ 
 \hline
 \multirow{2}{*}{5} & Proposed  & 0.693 & 0.015 & 0.048 \\ 
 \cline{2-5}
  & Conventional & 3.852 & 0.146 & 0.275 \\
\hline
 \multirow{2}{*}{10} & Proposed   & 1.912 & 0.031 & 0.057 \\ 
\cline{2-5}
  & Conventional  & 3.895 & 0.149 & 0.282 \\
\hline
 \multirow{2}{*}{15} & Proposed   & 5.445 & 0.081 & 0.117 \\ 
\cline{2-5}
  & Conventional  & 11.294 & 0.187 & 0.352 \\
\hline
 \multirow{2}{*}{20} & Proposed   & 8.158 & 0.125 & 0.158 \\ 
\cline{2-5}
  & Conventional  & 30.956 & 0.330 & 0.472 \\
\hline
\end{tabular}
\label{table:error_case4}
\end{table}

\subsection{Case 4: Solid mechanics problem}

We next demonstrate the proposed method on a more realistic engineering problem involving a finite element model from computational solid mechanics.

\subsubsection{Finite element forward model}

Consider a continuum body occupying the domain $\Omega \subset \mathbb{R}^3$, where $x\in\Omega$ denotes the spatial coordinate. The deformation of the body is described by the displacement field
\[
u(x) : \Omega \rightarrow \mathbb{R}^3 .
\]
Under the assumptions of small deformation and linear elasticity, the static equilibrium problem is to find the displacement field $u(x)$ satisfying
\begin{align}
\begin{cases}
\nabla_x \cdot \sigma_{\mathrm{stress}} = 0 & \text{in } \Omega,\\
u = \bar{u} & \text{on } \partial\Omega_u,\\
\sigma_{\mathrm{stress}} \cdot n = \bar{t} & \text{on } \partial\Omega_\sigma,
\end{cases}
\end{align}
where $\partial\Omega = \partial\Omega_u \cup \partial\Omega_\sigma$ and $\partial\Omega_u \cap \partial\Omega_\sigma = \varnothing$. 
Here $\bar{u}$ and $\bar{t}$ denote prescribed displacements and tractions, and $n$ is the outward unit normal.

The Cauchy stress tensor $\sigma_{\mathrm{stress}}$ is related to the infinitesimal strain tensor
\[
\varepsilon = \tfrac12(\nabla_x u + \nabla_x^{\top}u)
\]
through the constitutive relation
\[
\sigma_{\mathrm{stress}} = C : \varepsilon,
\]
where $C$ is the fourth-order isotropic elasticity tensor. For isotropic materials
\[
C = 3\kappa P_{\mathrm{vol}} + 2\mu P_{\mathrm{dev}},
\]
where $\kappa = \frac{E}{3(1-2\nu)}$ and $\mu = \frac{E}{2(1+\nu)}$ are the bulk and shear moduli, with $E$ and $\nu$ denoting Young’s modulus and Poisson’s ratio, respectively \citep{holzapfel2002nonlinear}.

Using the standard Galerkin finite element method \citep{hughes2012finite}, the weak form of the equilibrium equations is discretized on a conforming mesh. The resulting finite element system can be written as
\begin{align}
\label{e:FE_residual}
R(u) = Ku - G_{\mathrm{ext}} = 0,
\end{align}
where $u \in \mathbb{R}^{n_{\mathrm{dof}}}$ is the vector of nodal displacements and $n_{\mathrm{dof}}$ denotes the number of displacement degrees of freedom.

The global stiffness matrix is assembled from element contributions
\[
K = \mathcal{A}_{e=1}^{n_{\mathrm{ele}}} K^e,
\qquad
K^e = \int_{\Omega^e} B^{\top}[C]B\,\mathrm{d}v,
\]
while the external force vector is
\[
G_{\mathrm{ext}} =
\mathcal{A}_{e\in S_\sigma} G_{\mathrm{ext}}^e,
\qquad
G_{\mathrm{ext}}^e =
\int_{\partial\Omega_\sigma^e} N^{\top}t\,\mathrm{d}s,
\]
where $N$ and $B$ denote the shape function and strain-displacement matrices, respectively.

\subsubsection{Inference and prediction setup}

\begin{figure}[htbp]
\centering
\includegraphics[scale=0.7]{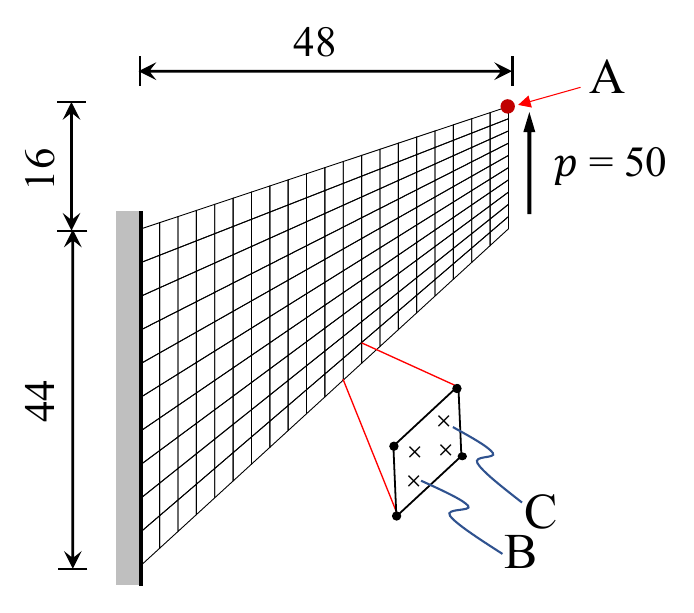}
\caption{Case 4: Geometry and finite element mesh of the Cook's membrane problem.}
\label{fig:mesh}
\end{figure}

We consider the classical Cook's membrane benchmark problem, whose geometry and finite element mesh are shown in Figure~\ref{fig:mesh}. The mesh consists of 200 four-node quadrilateral elements.

The uncertain parameters in this problem are Young's modulus $E$ and Poisson's ratio $\nu$. 
To ensure physical constraints on these parameters, we introduce latent variables
\[
\theta =
\begin{bmatrix}
\theta_1 \\
\theta_2
\end{bmatrix},
\qquad
\theta_1,\theta_2 \sim \mathcal{N}(0,1),
\]
and define the parameter transformation
\begin{align}
g(\theta)
=
\begin{bmatrix}
E \\
\nu
\end{bmatrix}
=
\begin{bmatrix}
e^{\theta_1} \\
\dfrac{0.5}{1+e^{-\theta_2}}
\end{bmatrix}.
\end{align}
This construction ensures that $E>0$ and $\nu\in(0,0.5)$, which are the physically admissible ranges for isotropic elastic materials.

\paragraph{Observation model}

The observable quantities are the horizontal and vertical displacements at point $A$ in the structure (see Figure~\ref{fig:mesh}). Let $u_A$ denote these two displacement components. The observation model is defined as
\begin{align}
y = u_A = G_{\mathrm{FEM}}(E,\nu) + \epsilon,
\end{align}
where $\epsilon \sim \mathcal{N}(0,\sigma_\epsilon^2=10^{-1})$ represents measurement noise.
The mapping $G_{\mathrm{FEM}}(E,\nu)$ corresponds to solving the finite element equilibrium equations in \eqref{e:FE_residual} and extracting the displacement components at node $A$:
\begin{align}
G_{\mathrm{FEM}}(E,\nu) = I_A u,
\end{align}
where $I_A \in \mathbb{R}^{2 \times n_{\mathrm{dof}}}$ is a selection matrix that extracts the horizontal and vertical displacement degrees of freedom at node $A$.

\paragraph{Prediction model}

The predictive QoI in this example is the von Mises stress at two integration points $B$ and $C$ located within a selected element (see Figure~\ref{fig:mesh}). The predictive model is defined as
\begin{align}
z = \sigma_{BC} = H_{\mathrm{FEM}}(E,\nu) + \eta,
\label{eq50}
\end{align}
where $\eta \sim \mathcal{N}(0,\sigma_\eta^2= 3\times10^{-3})$ represents predictive noise.
The mapping $H_{\mathrm{FEM}}(E,\nu)$ computes the von Mises stresses at the two integration points:
\begin{align}
H_{\mathrm{FEM}}(E,\nu) =
\begin{bmatrix}
\sqrt{\tfrac32\, s(x_B):s(x_B)} \\
\sqrt{\tfrac32\, s(x_C):s(x_C)}
\end{bmatrix},
\end{align}
where $s = P_{\mathrm{dev}}:\sigma_{\mathrm{stress}}$ denotes the deviatoric part of the Cauchy stress tensor.

\medskip
In summary, the observation variable is the displacement vector $y=u_A$, while the predictive QoI is the stress vector $z=\sigma_{BC}$.

\subsubsection{Results}

Due to the higher computational cost for the finite element forward solves, the conventional approach now uses $N_c=10^4$. For the reference solution, $10^4$ MCMC samples are generated to produce the same number of von Mises stress samples.

To illustrate the offline--online computational tradeoff, we report in Table~\ref{table:CPU_time} the online time required to predict $p(z|y)$ for a representative observation $y=[0.1,0.1]^\top$. This timing study was performed on a MacBook Pro with an Apple M2 Max chip, 12 CPU cores, and 38 GPU cores. The results show that the proposed method has substantially lower online prediction time than the conventional variational and reference methods. This speedup reflects the amortized formulation: the computationally intensive training has already been performed offline.

For this online query, the proposed method is more than two orders of magnitude faster than the conventional variational method and more than three orders of magnitude faster than the reference method. This comparison concerns online prediction time only; the proposed method also requires an offline training stage, which took 111617.2 seconds in this example.

\begin{table}[htbp]
\caption{Case 4: Computational time (in seconds) for predicting $p(z|y)$ with $y=[ 0.1, 0.1]^{\top}$ from proposed, conventional, and reference methods.}
\centering
\begin{tabular}{|c|c|c|} 
 \hline
 \textbf{Proposed method} & \textbf{Conventional method} & \textbf{Reference} \\ 
 \hline
 0.2 & 86.0 & 281.8 \\
 \hline
\end{tabular}
\label{table:CPU_time}
\end{table}

\begin{figure}[htbp]
 \includegraphics[scale=0.48]{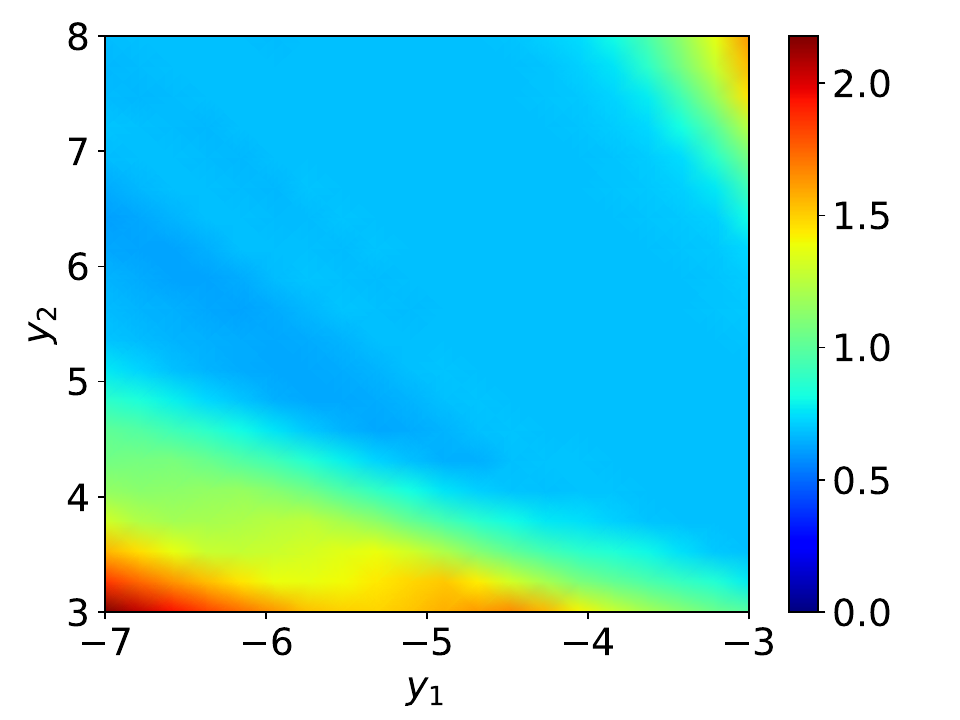}
 \includegraphics[scale=0.48]{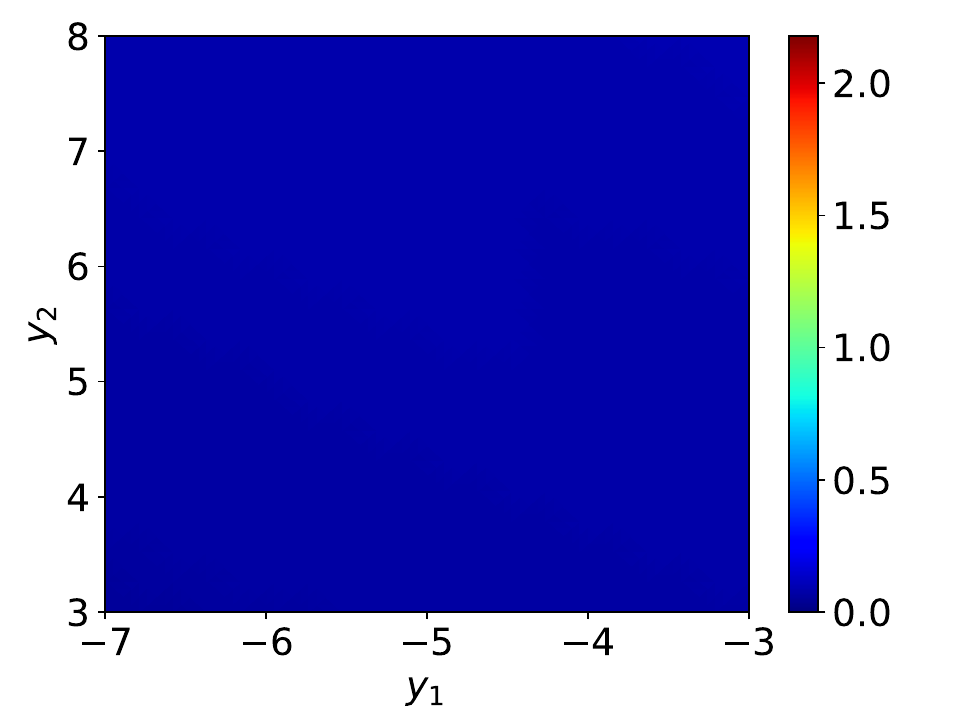}
\caption{Case 4: KL divergence between the approximate and reference posterior-predictive distributions for the (left) conventional and (right) proposed methods.}
\label{fig:kld_case4}
\end{figure}

\begin{figure}[htbp]
\includegraphics[scale=0.6]{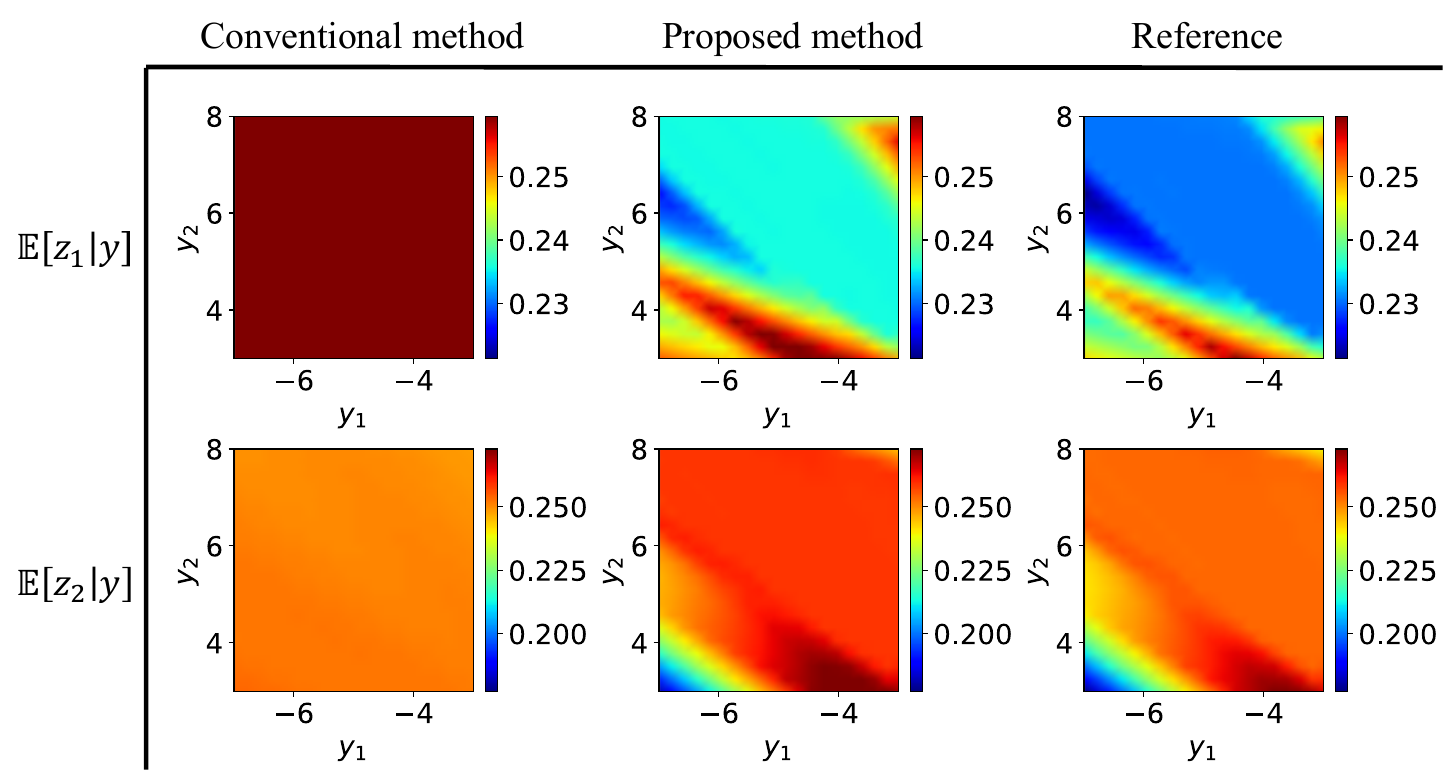}
\centering
\caption{Case 4: Mean of the posterior-predictive distributions. Rows correspond to the two components $z_1$ and $z_2$, and columns show the conventional, proposed, and reference solutions.}
\label{fig:mean_case4}
\end{figure}

\begin{figure}[htbp]
\includegraphics[scale=0.6]{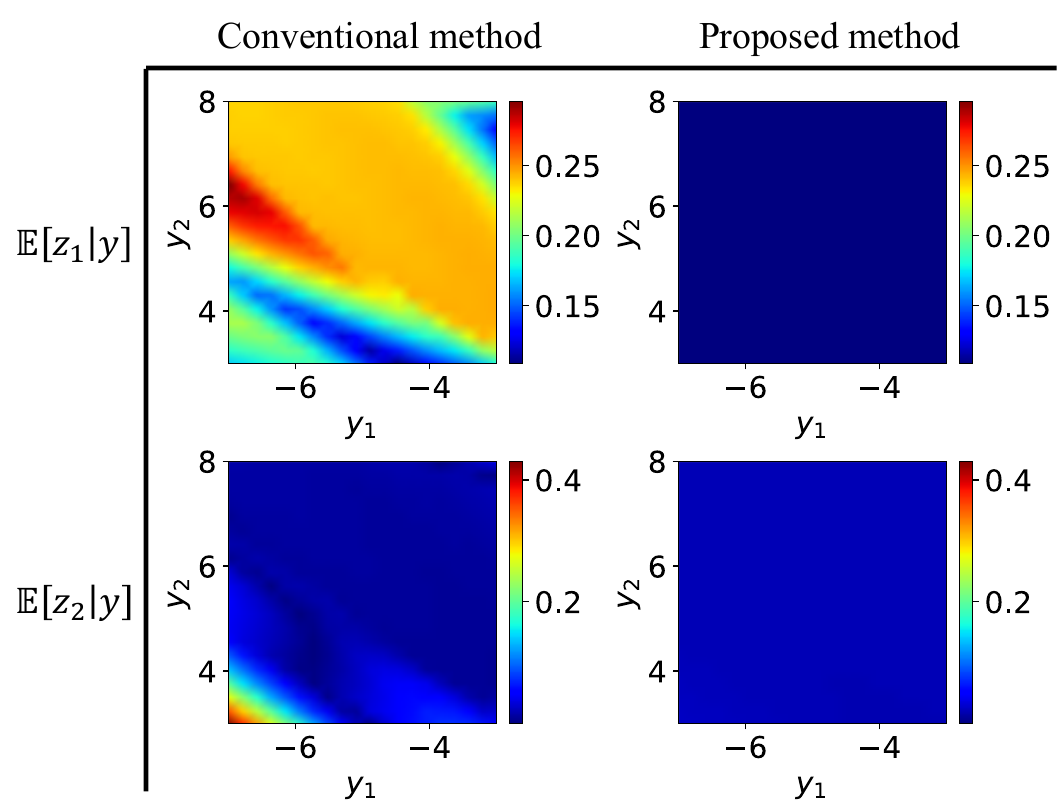}
\centering
\caption{Case 4: Relative errors of the posterior-predictive mean. Rows correspond to the two components $z_1$ and $z_2$, and columns show the conventional and proposed methods.}
\label{fig:rela_mean_case4}
\end{figure}

\begin{figure}[htbp]
\includegraphics[scale=0.6]{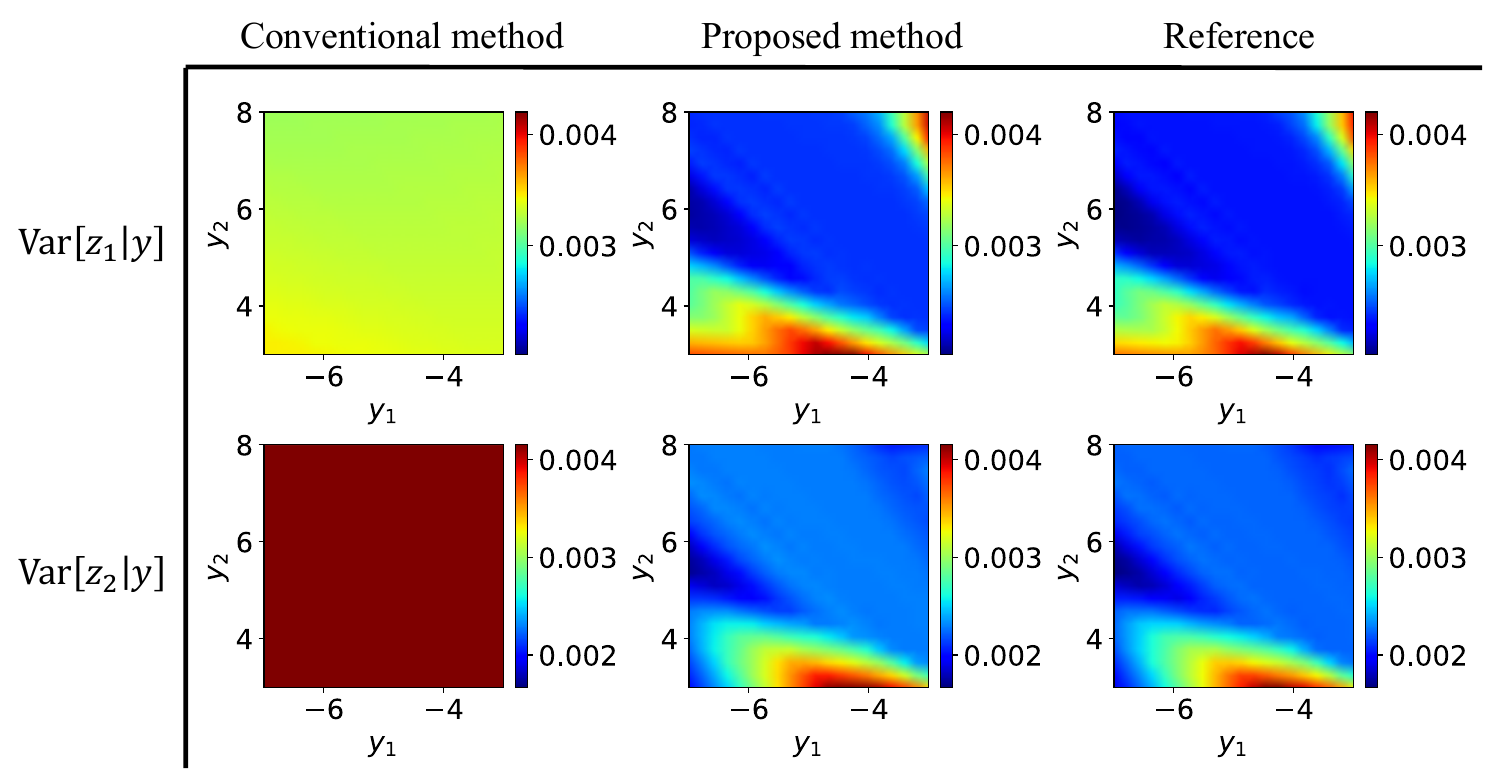}
\centering
\caption{Case 4: Variance of the posterior-predictive distributions. Rows correspond to the two components $z_1$ and $z_2$, and columns show the conventional, proposed, and reference solutions.}
\label{fig:sig_case4}
\end{figure}

\begin{figure}[htbp]
\includegraphics[scale=0.6]{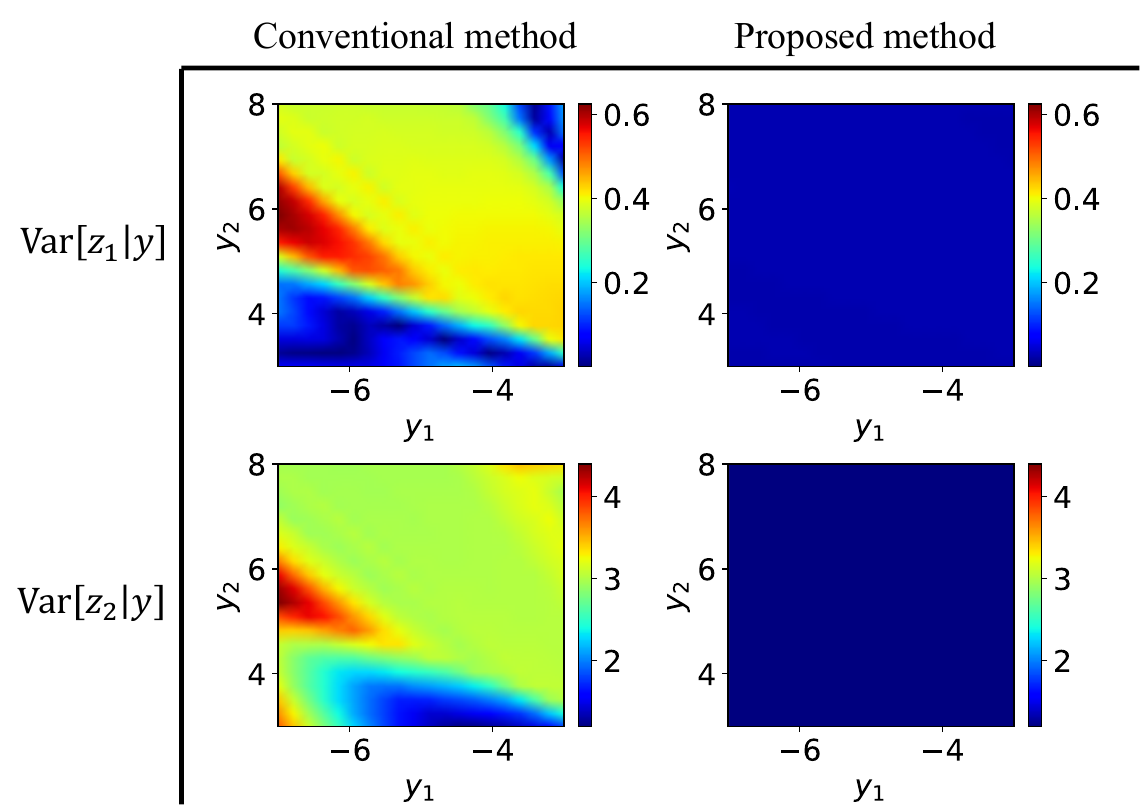}
\centering
\caption{Case 4: Relative errors of the posterior-predictive variance. Rows correspond to the two components $z_1$ and $z_2$, and columns show the conventional and proposed methods.}
\label{fig:rela_sig_case4}
\end{figure}

\begin{figure}[htbp]
\includegraphics[scale=0.65]{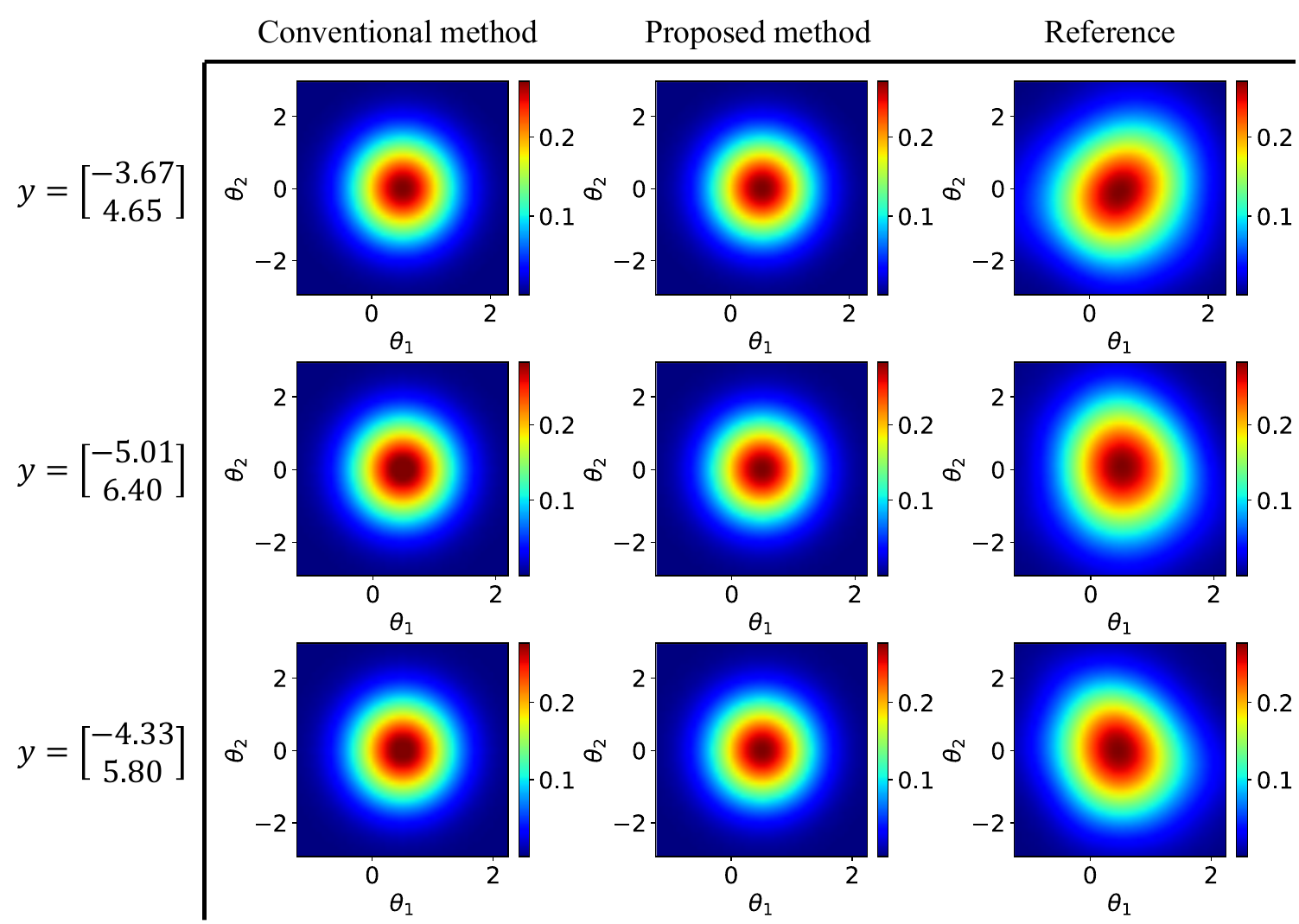}
\centering
\caption{Case 4: Examples of posterior distributions for different values of $y$. Rows correspond to the two components $\theta_1$ and $\theta_2$, and columns show the conventional, proposed, and reference methods.}
\label{fig:p_theta_y_case4}
\end{figure}

\begin{figure}[htbp]
\includegraphics[scale=0.65]{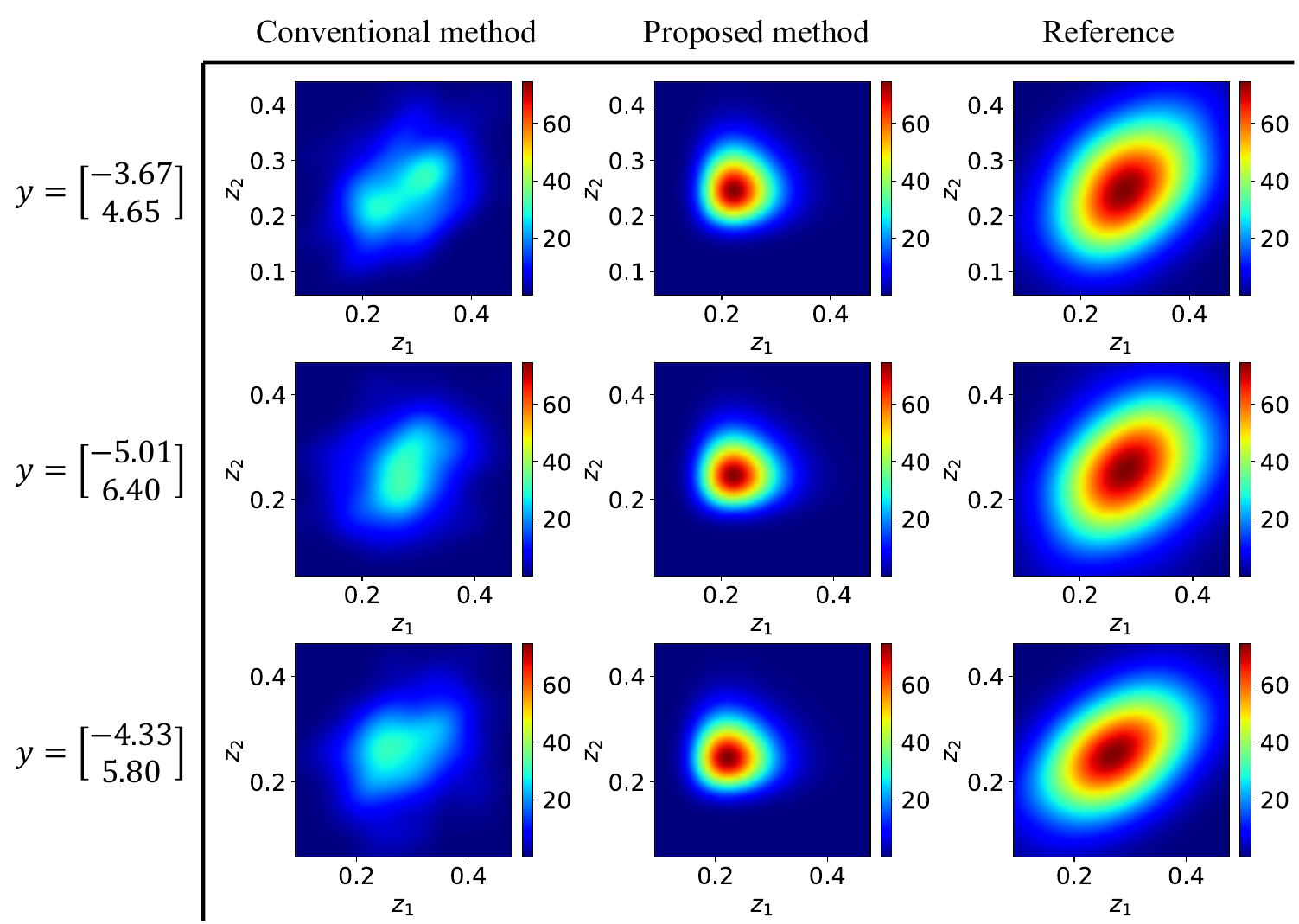}
\centering
\caption{Case 4: Examples of posterior-predictive distributions for different values of $y$. Rows correspond to the two components $z_1$ and $z_2$, and columns show the conventional, proposed, and reference methods.}
\label{fig:p_z_y_case4}
\end{figure}

Figure \ref{fig:kld_case4} highlights the proposed method's superior accuracy, revealing a consistently lower KL divergence between its estimated posterior-predictive distribution and the reference solution compared to the conventional approach. This indicates a significantly better distributional fit. The improved accuracy is further confirmed by the distribution's moments, where the proposed method's estimated mean and variance align more closely with the reference solutions (Figures \ref{fig:mean_case4} and \ref{fig:sig_case4}). Consequently, the relative errors for both the mean and variance are consistently smaller for the proposed method across the varying measurements (Figures \ref{fig:rela_mean_case4} and \ref{fig:rela_sig_case4}). 
Additionally, the proposed method achieves posterior estimates comparable to those of the conventional method across different observations (Figure~\ref{fig:p_theta_y_case4}). For the posterior-predictive distribution, the proposed method accurately captures the regions of high probability density in the reference solutions, whereas the conventional method fails to reproduce these features (Figure~\ref{fig:p_z_y_case4}).

\section{Conclusions}
\label{conclusions}

This work presents a VI framework for approximating posterior and posterior-predictive distributions in Bayesian UQ. 
In contrast to the conventional two-stage approach that first infers the posterior distribution of model parameters and then propagates samples through a predictive model, the proposed method directly learns variational approximations of both the posterior and posterior-predictive distributions through a unified optimization problem. 
The framework incorporates a variational upper bound on the predictive KL divergence together with moment-based regularization, in order to achieve stable and accurate learning of predictive distributions.

The proposed method accommodates different variational parameterizations.
Independent Gaussian variational distributions are used when predictive QoIs are unbounded, while log-normal predictive distributions are employed when physical quantities are constrained to be positive. 
The resulting variational distributions are trained in an amortized manner, allowing the computationally intensive training phase to be performed offline. 
After training, predictive inference can be obtained through a simple evaluation of the learned variational distributions.

The performance of the proposed framework is demonstrated through a series of numerical examples of increasing complexity, including linear and nonlinear benchmark problems and a finite-element solid mechanics model governed by a PDE. 
Across all examples, the proposed method consistently produces more accurate approximations of the posterior-predictive distribution than the conventional two-stage VI approach, as measured by KL divergence and moment-based error metrics. 
Furthermore, the proposed method enables substantially reduced online computational cost for predictive inference by shifting computation to an offline training stage.

Despite these promising results, several limitations remain. 
First, while the method is demonstrated in linear settings across a range of dimensions, its performance in nonlinear, high-dimensional parameter and observation spaces requires further investigation. 
Second, the tightness of the variational upper bound used in the predictive objective remains an open question and warrants additional theoretical analysis. 
Third, the current variational parameterizations, based primarily on independent Gaussian and log-normal distributions, may struggle to capture strongly multi-modal predictive distributions.

Future work will focus on extending the framework to more expressive variational families capable of representing complex and higher-dimensional distributions. 
Potential directions include transport maps~\cite{ElMoselhy2012,Marzouk_16_Transport,Wang2025} and normalizing flows~\cite{Kobyzev_20_NFsReview,papamakarios2021normalizing}. 
These approaches offer promising avenues for improving the flexibility and accuracy of predictive UQ often needed in complex scientific and engineering applications.

\section*{Acknowledgments}

This research is supported in part through computational resources and services provided by the Google Cloud Research Credits program and Advanced Research Computing at the University of Michigan, Ann Arbor.

\bibliographystyle{elsarticle-num} 
\bibliography{refs} 

\end{document}

%% file: macros.tex
\newcommand{\argmin}{\operatornamewithlimits{argmin}}
\newcommand{\DKL}{D_{\mathrm{KL}}}